\documentclass[10pt, letter, onecolumn]{arxiv}

\usepackage{kantlipsum, lipsum}
\usepackage{dm-colors}
\usepackage{amsmath}
\usepackage{pstricks, pst-node}
\usepackage{verbatim}
\usepackage{multirow}
\usepackage{scalerel}
\usepackage{booktabs}
\usepackage{enumitem}
\usepackage{xspace}
\usepackage{bm}
\usepackage{bbm}
\usepackage{mathtools}
\usepackage{soul}
\usepackage{epsfig}
\usepackage{graphicx}
\usepackage{csquotes}
\usepackage{setspace}
\usepackage{colortbl}
\usepackage{threeparttable}
\usepackage{tabularx,ragged2e}
\usepackage{placeins}
\usepackage[hang,flushmargin,symbol]{footmisc}

\usepackage{varioref}
\usepackage[pagebackref=false,breaklinks=false,%
            colorlinks=true,bookmarks=true,citecolor=ourdarkblue,%
            urlcolor=ourdarkblue,linkcolor=ourdarkblue]{hyperref}
\usepackage[noabbrev,capitalize]{cleveref}
\usepackage{etoc}
\usepackage{lineno}

\graphicspath{{figures/}}

\newcommand{\change}[1]{{\textcolor{black}{#1}}}

\usepackage[misc]{ifsym}

\title{\Large{Towards Generalist Foundation Model for Radiology by Leveraging Web-scale 2D\&3D Medical Data}}

\vspace{1cm}
\author[$\ast$,1,2]{Chaoyi Wu} 
\author[$\ast$,1,2]{Xiaoman Zhang}
\author[1,2]{\\ \vspace{0.1cm} Ya Zhang}
\author[1,2, \Letter]{Yanfeng Wang} 
\author[1,2, \Letter]{Weidi Xie}

\affil[1]{\normalsize Shanghai Jiao Tong University \hspace{1cm}}

\affil[2]{\normalsize Shanghai AI Laboratory \authorcr 

}

\renewcommand{\correspondingauthor}[1]{$\ast$~Equal contributions. Email addresses: \{wtzxxxwcy02, xm99sjtu, weidi\}@sjtu.edu.cn }

\begin{document}

\begin{abstract}

In this study, we aim to initiate the development of 
\textbf{Rad}iology \textbf{F}oundation \textbf{M}odel, termed as \textbf{RadFM}. 
\change{We consider the construction of foundational models from three perspectives, 
namely, dataset construction, model design, and thorough evaluation.}
Our contribution can be concluded as follows: 
(\textit{i}), 
we construct a large-scale \textbf{Med}ical \textbf{M}ulti-modal \textbf{D}ataset, 
{\bf MedMD}, which consists of 16M 2D and 3D medical scans with high-quality text descriptions or reports across various data formats, modalities, and tasks, covering over 5000 distinct diseases.
To the best of our knowledge, this is the first large-scale, 
high-quality, medical visual-language dataset, with both 2D and 3D scans; 
(\textit{ii}), we propose an architecture that enables visually conditioned generative pre-training, {\em i.e.}, allowing for integration of text input with 2D or 3D medical scans, 
and generate responses for diverse radiologic tasks. The model was initially pre-trained on MedMD and subsequently fine-tuned on the domain-specific dataset, which is a radiologic cleaned version of MedMD, containing 3M radiologic visual-language pairs, termed as RadMD;  
(\textit{iii}), we propose a new evaluation benchmark, \textbf{RadBench}, 
that comprises five tasks, including modality recognition, disease diagnosis, visual question answering, report generation and rationale diagnosis, aiming to comprehensively assess the 
capability of foundation models in handling practical clinical problems.
We conduct both automatic and human evaluation on RadBench, 
\change{in both cases, RadFM outperforms existing multi-modal foundation models, that are publicaly accessible, including Openflamingo, MedFlamingo, MedVInT and GPT-4V. 
Additionally, we also adapt RadFM for different public benchmarks, 
surpassing existing SOTAs on diverse datasets.} 
All codes, data, and model checkpoint will all be made publicly available to promote further research and development in the field.
\vspace{0.5cm}



\end{abstract}

\maketitle

\tableofcontents

\clearpage
\section{Introduction}

Generalist foundation models~\cite{bommasani2021opportunities}, 
the latest generation of AI models pre-trained on large-scale dataset, 
have demonstrated remarkable success in various domains, for example, 
natural language processing, computer vision~\cite{touvron2023llama, Li2023BLIP2BL}. 
Their ability to address diverse and challenging tasks has also attracted tremendous attention among researchers in the field of Artificial Intelligence for Medicine~(AI4Medicine)~\cite{li2023llava, Moor2023FoundationMF, moor2023medflamingo, tu2023towards, Zhang2023PMCVQAVI}. Despite the promising clinical use cases, the progress in developing medical foundation models has been fundamentally hindered by three challenges:

\vspace{-0.15cm}
\begin{itemize}
\setlength\itemsep{0.15cm}
\item {\bf Lack of multimodal datasets for training:} medicine by its nature, requires understanding multimodal data, spanning text~(electronic health record, medical reports), 1D signals~(ECG), 2D images~(ultrasound, X-ray), 3D images~(CT or MRI scans), genomics, and more. To support the training of medical generalist foundation model,
a large-scale, diverse, multimodal dataset is desperately required; 

\item {\bf Lack of general architecture formulation:} 

in the literature of AI4Medicine, various clinical tasks have largely been tackled by following a divide-and-conquer paradigm, 
{\em i.e.}, different architectures are designed for the problem of interest, 
like diagnosis~\cite{tiu2022expert, zhang2023knowledge} or 
\change{
report generation~\cite{monshi2020deep, yu2023evaluating}.}
In contrast, developing a medical foundation model requires one general architecture that is capable of tackling a wide spectrum of clinical tasks, by fusing information from a mixture of different modalities;

\item {\bf Lack of effective benchmark to monitor progress:} 
benchmarking the models' clinical knowledge predominantly relies on task-specific datasets with a limited number of testing cases. An high-quality benchmark is yet to be established, 
to comprehensively measure the progress of the development on medical foundation model across a wide range of clinical tasks.
\end{itemize}


\vspace{-0.15cm}
Considering the above-mentioned challenges, in this paper, we take a preliminary, yet realistic step toward developing a generalist medical foundation model for radiology, which has shown to play a vital role in clinical scenarios, for example, disease diagnosis, treatment planning, and monitoring patient progression. \change{Specifically, we present our progress towards building a \textbf{Rad}iology \textbf{F}oundation \textbf{M}odel~(\textbf{RadFM}), that aims to tackle a wide spectrum of clinical radiology tasks, by learning from medical scans (X-ray, CT, MRI, PET, {\em etc.}) and corresponding text description/reports.}

\begin{figure}[t]
    \centering
    \includegraphics[width = \textwidth]{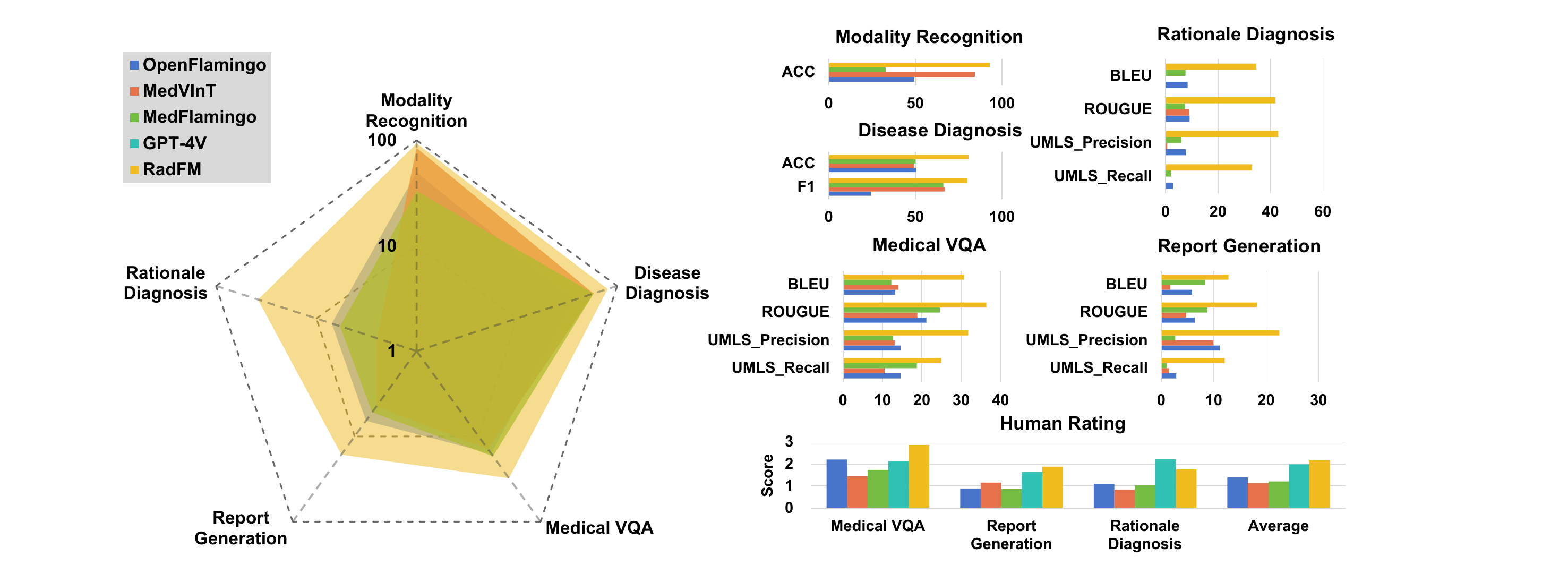}
    \vspace{2pt}
    \caption{
    \change{The general comparison between RadFM and different SOTA methods, \emph{i.e.}, OpenFlamingo~\cite{Alayrac2022FlamingoAV}, MedVInT~\cite{Zhang2023PMCVQAVI}, Med-Flamingo~\cite{moor2023medflamingo} and GPT-4V~\cite{GPT4V}. 
    On the left we plot the radar figure of the five models, 
    on the average of different automatic metrics, 
    the coordinate axes are \textbf{logarithmized}. 
    On the top right, we draw comparison on four different tasks with different automatic metrics in detail. 
    On the bottom right, we show the human rating comparison of the five models under three open-ended task types~(medical VQA, report generation, and rationale diagnosis). 
    All evaluations have shown the superiority of RadFM, surpassing existing methods by a noticable margin.}}
    \label{fig:teaser}
    \vspace{-0.6cm}
\end{figure}

\change{To achieve this, we start by constructing a novel, large-scale, 
\textbf{Med}ical \textbf{M}ulti-modal \textbf{D}ataset, named \textbf{MedMD},
consisting of \textbf{16M} 2D and 3D radiology scans, accompanied with high-quality textual descriptions, for example, radiology reports, visual-language instruction, or crucial disease diagnosis labels. 
MedMD encompasses a wide range of radiological modalities,
covering \textbf{17} medical systems, {\em e.g.}, Breast, Cardiac, Central Nervous System, Chest, Gastrointestinal, Gynecology, Hematology, Head and Neck, Hepatobiliary, Musculoskeletal, Obstetrics, Oncology, Pediatrics, Spine, Trauma, Urogenital and Vascular featuring over {\bf 5000} diseases, thus potentially serving as the cornerstone for developing foundation models in radiology.} 


Architecturally, RadFM refers to a visually conditioned autoregressive text generation model, 
that enables to seamlessly integrate natural language with 2D or 3D medical scans, 
and address a wide range of medical tasks with natural language as output. 
The proposed model is initially pre-trained on the large \textbf{MedMD} dataset, 
and subsequently fine-tuned via visual instruction on a filtered subset, 
comprising \textbf{3M} meticulously curated multi-modal samples with only radiologic cases, 
termed as \textbf{RadMD}, ensuring a high-quality and reliable dataset for the domain-specific fine-tuning process.

To monitor the developmental progress of the foundation model for radiology, 
we establish a novel, comprehensive evaluation benchmark, \textbf{RadBench},
covering a variety of clinical tasks, for example, disease diagnosis, report generation, and visual question-answering on radiologic modalities and anatomical regions. 
All samples in RadBench have undergone meticulous manual verification to ensure data quality.
\change{
We conduct both automatic and human evaluation on RadBench with existing strong models that are publicaly accessible, 
for example, Open-flamingo~\cite{anas_awadalla_2023_7733589}, MedVInT~\cite{Zhang2023PMCVQAVI}, MedFlamingo~\cite{moor2023medflamingo} and GPT-4V~\cite{GPT4V}, and observe significant benefits across all considered tasks. 
In addition, we perform adaptation of RadFM on several public benchmarks, demonstrating the generalization ability of RadFM.}

\change{
Overall, in this work, we demonstrate the potential of developing a generalist foundation model for radiology, by making contributions from three key aspects: a large-scale multimodal radiology dataset (RadMD), a demonstration of the radiology foundation model (RadFM), and a comprehensive benchmark for radiology to monitor progress~(RadBench).}

\begin{figure}[t]
    \centering
    \includegraphics[width = \textwidth]{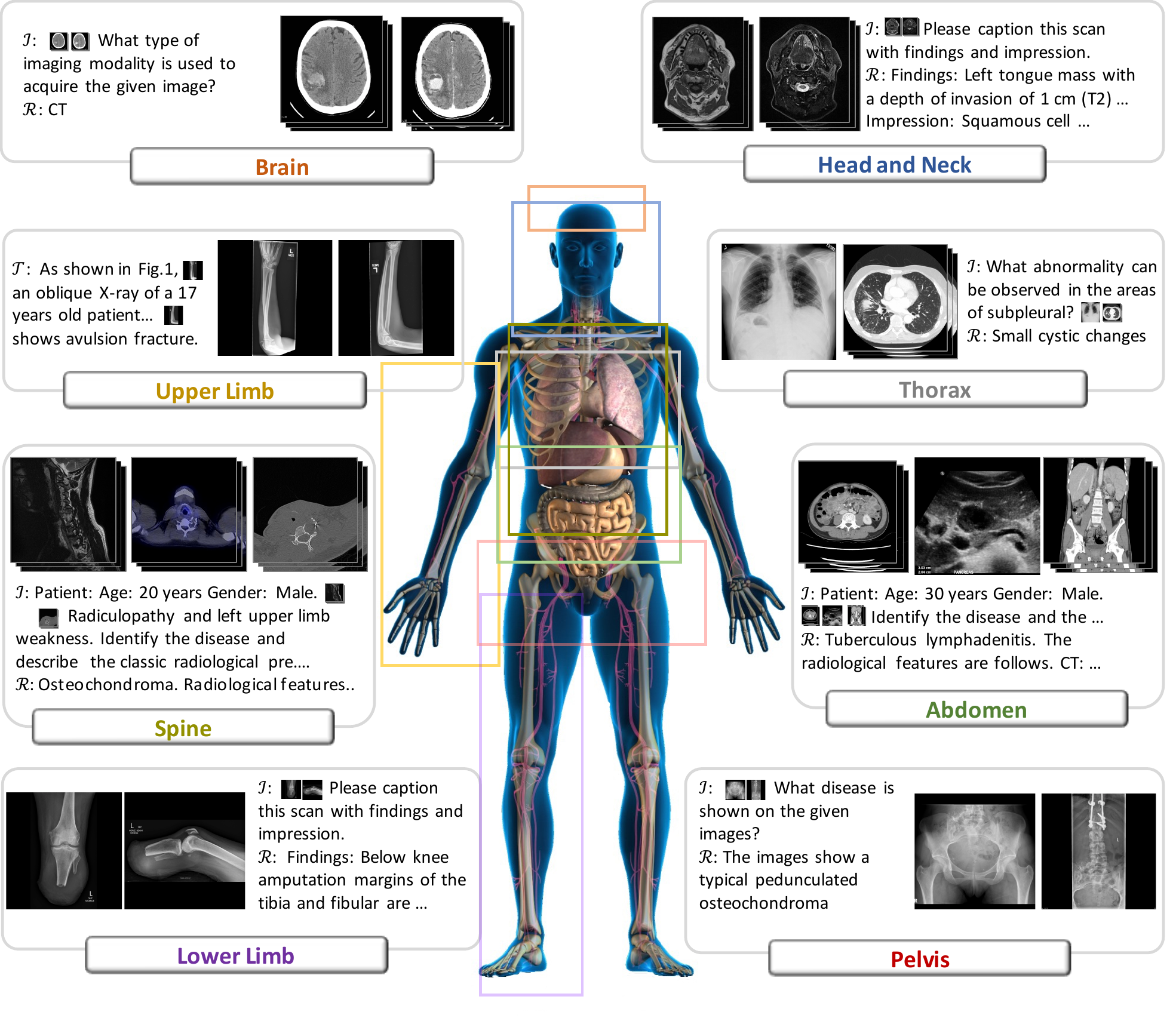}
    \caption{\change{Overview of Medical Multimodal Dataset~(MedMD). Our collected data covers the majority of radiologic modalities and anatomical regions of the human body, such as brain, head and neck, thorax, spine, abdomen, upper limb, lower limb, and pelvis, etc. 
    The dataset mixes two types of datasets, {\em i.e.}, interleaved datasets and visual instruction datasets. $\mathcal{T}$ refers to the text of interleaved data, $\mathcal{I}$ refers to the instruction input text, and $\mathcal{R}$ refers to the response text. }
    }
    \label{fig:data_view}
\end{figure}

\section{Dataset}
\change{Here, we describe the procedure for constructing the datasets and benchmark. In Sec.~\ref{sec:medmd}, we present {\bf Med}ical {\bf M}ultimodal {\bf D}ataset~({\bf MedMD}) along with a filtered radiology subset {\bf Rad}iology {\bf M}ultimodal {\bf D}ataset~({\bf RadMD}). To the best of our knowledge, MedMD is the first large-scale, high-quality medical vision-language dataset, covering a wide range of anatomies with over 5000 diseases. 
}
In Sec.~\ref{sec:radbench}, we introduce a new \textbf{Rad}iology \textbf{Bench}mark for evaluation, termed as \textbf{RadBench}, with five distinct tasks, 
\emph{e.g.}, modality recognition, disease diagnosis, visual question
answering, report generation and rationale diagnosis, 
aiming to monitor the progress of developing foundation models.

\subsection{Medical Multimodal Dataset~(MedMD)}
\label{sec:medmd}
To start, we construct a candidate data pool by pulling a variety of existing visual-language medical datasets together, for example, MIMIC-CXR~\cite{johnson2019mimic} and PMC-OA~\cite{lin2023pmc}. 
Despite the scale of these high-quality datasets, they are fundamentally limited in several aspects:
{\bf (i) Data format.} These datasets are only composed of 2D medical images, 
which do not fully capture the complexities in clinical use cases, 
for example, 3D medical imaging modalities, like CT, MRI;
{\bf (ii) Modality diversity.} A noteworthy limitation arises from the fact only chest X-ray images are provided with medical reports, \change{training models on such data will clearly pose limitation on the generalizability to a broader range of imaging modalities and anatomical regions}; 
{\bf (iii) Report quality}. Another critical limitation lies in the use of data extracted from figures and captions from research papers. The gap between research-oriented data and real-world clinical scenarios may not support accurate and reliable clinical diagnoses.
Therefore, to support the training of our proposed Radiology Foundation Model~(RadFM), 
\change{
we augment the dataset with four new ones, 
including PMC-Inline, PMC-CaseReport, RP3D-Series, and MPx-Series, resulting in MedMD.
MedMD has a total of 16M 2D image-text pairs, including 15.5M 2D images and 500k 3D scans with corresponding captions or diagnosis labels, as shown in Tab.~\ref{dataset}.}

Generally speaking, we split the candidate data pool into two parts, 
\change{(i) interleaved image-language data that is collected from academic papers, 
(ii) image-language data constructed for visual-language instruction tuning,} as detailed below.

\begin{table}[!thb]
\footnotesize
\centering
\renewcommand{\arraystretch}{1.2}
\caption{
\change{Description of the collected dataset {\bf Med}ical {\bf M}ultimodal {\bf D}ataset~(MedMD) for model pre-training and the filtered dataset {\bf Rad}iology {\bf M}ultimodal {\bf D}ataset~(RadMD) for domain-specific fine-tuning. 
\textbf{Filter Strategy} refers the procedure to curate RadMD.
\textbf{Size} refers to the pair size for both image and image-text data.}
}
\vspace{5pt}
\begin{tabular}{lp{7.5cm}<{\raggedright}lll}
\toprule
\rowcolor{lightgray}
\textbf{Dataset Name} & \textbf{Description} & \textbf{Filter Strategy} & \textbf{Size} & \textbf{Filter Size}\\ 
\midrule
\rowcolor{lightgray!50}
\multicolumn{5}{l}{\textbf{Interleaved Dataset } } \\
PMC-Inline & A medical dataset containing PMC-papers which links with images through inline reference, \emph{e.g.}, ``as shown in fig.X''. & Filtered out & 11M  & 0 \\
\midrule
\rowcolor{lightgray!50}
\multicolumn{5}{l}{\textbf{Visual Instruction Tuning Dataset} } \\
\rowcolor{lightgray!20}
Image data & & & & \\
VinDr-Mammo~\cite{vindrmammo} & A mammography dataset consists of four-view exams with level assessment and finding annotations. & Keep all & 50k & 50k \\
VinDr-SpineXR~\cite{vindrspine} & A dataset consists of spine X-ray images with annotations of 13 types of abnormalities. & Keep all & 
14k & 14k \\
VinDr-PCXR~\cite{vindrpcxr} & A dataset consists of pediatric chest X-ray images with annotations of 36 critical findings and 15 diseases. & Keep all &  6.5k & 6.5k  \\
CXR-Mix~\cite{Wu2023KDiagKD} & A collection of different chest X-ray diagnosis datasets. & Keep all & 668k & 668k \\
RadChest-CT~\cite{draelos_rachel_lea_2020_6406114} & A dataset consists of chest CT scans labeled with 84 abnormality labels and 52 location labels. & Keep all & 73k & 73k \\
\rowcolor{lightgray!20}
\multicolumn{5}{l}{2D Image-text data } \\
PMC-OA~\cite{lin2023pmc} & A medical dataset contains paired figures and captions collected from PubMed Central. & Filtered out & 1.65M & 0\\  
PMC-VQA~\cite{Zhang2023PMCVQAVI} & A medical visual question-answering dataset generated from PMC-OA. & Keep all & 413k & 413k\\ 
PMC-CaseReport & A sub-dataset filtered from PMC-Inline containing 103K cases report papers. We generate VQA pairs by querying ChatGPT. we keep some background information of the patient to form context input. & Keep radiology &  1.1M & 438k \\
MPx-Single & A medical vision-language dataset contains the modality, plane and captions for each image.& Keep all & 120k & 120k\\
MPx-Multi & A medical vision-language dataset contains findings, discussion, and diagnoses for each case which may contain a series of radiology images. & Keep all & 39K & 39K\\
VQA-RAD~\cite{lau2018dataset} & A medical visual question-answering dataset consists of question–answer pairs on 315 radiology images.& Keep all & 3.5K & 3.5K \\
SLAKE~\cite{liu2021slake} & A bilingual visual question-answering dataset consisting of 642 images. & Keep all
& 6K & 6K \\
MIMIC-CXR~\cite{johnson2019mimic} & A chest image-report dataset contains 377k images corresponding to 227k studies. & Keep all & 227K & 227K\\
\rowcolor{lightgray!20}
\multicolumn{5}{l}{2D $\&$ 3D Image-text data } \\
RP3D-Caption & A medical dataset consists of images and corresponding captions.& Keep radiology &  73k & 51k\\
RP3D-VQA & A medical visual question-answering dataset generated from the captions in RP3D-Caption. & Keep radiology  & 205k & 142k \\
RP3D-Modality & A medical vision-language dataset contains the modality question for each image. & Keep radiology & 264K & 236k \\ 
RP3D-Rationale &  A medical vision-language dataset contains disease rationale diagnosis for each case which may contain a series of radiology images. & Keep radiology & 73K & 43k \\
\bottomrule
\end{tabular}
\label{dataset}
\end{table}

\subsubsection{Interleaved Dataset}
\vspace{3pt} \noindent \textbf{PMC-Inline.} 
PMC-Inline contains 11M 2D radiology images that are collected from PubMed Central papers. In contrast to existing work, for example, PMC-OA~\cite{lin2023pmc}, that only contains figures and corresponding captions, here, we focus on the inline reference from the main body of papers. 
For example, one paper may contain many sentences like ``As shown in Fig.2, we can see \dots'', we localise the keyword ``Fig.2'' and link its corresponding figure back into sentences, 
ending up with interleaved images and texts, with rich context. 
This dataset shares the same format as MMC4~\cite{zhu2023multimodal}, 
which has shown to be effective in training foundation models in computer vision community, 
for example, Flamingo~\cite{Alayrac2022FlamingoAV}.



\subsubsection{Visual-language Instruction Tuning Dataset}

\vspace{3pt} \noindent \textbf{PMC-CaseReport.} 
PMC-CaseReport is a filtered subset of PMC-Inline with around 103K case reports,
where the doctors typically document the valuable clinical cases, 
based on their contact with the patients, such as family medical history, preliminary diagnosis, radiographic exam results, surgical records, {\em etc.}, 
together with critical radiologic scans, that generally follows the real timeline. 

Similar to PMC-VQA~\cite{Zhang2023PMCVQAVI} that generates VQA pairs by querying ChatGPT with image captions, we also generate 1.1M question-answer pairs by querying ChatGPT with the sentences containing inline references in case reports. However, in contrast to PMC-VQA, we keep background information of the patients to simulate the clinical diagnosis scenario, thus can be seen as a medical contextual VQA dataset. For example, a question-answer pair may like ``Question: A 58-year-old woman presented to the emergency department \dots Postoperatively, her pain significantly relieved. What did the MRI indicate? Answer: The MRI indicated tumor recurrence at L2 and S1-S2.''

\begin{figure}[!htb]
    \centering
    \includegraphics[width = \textwidth]{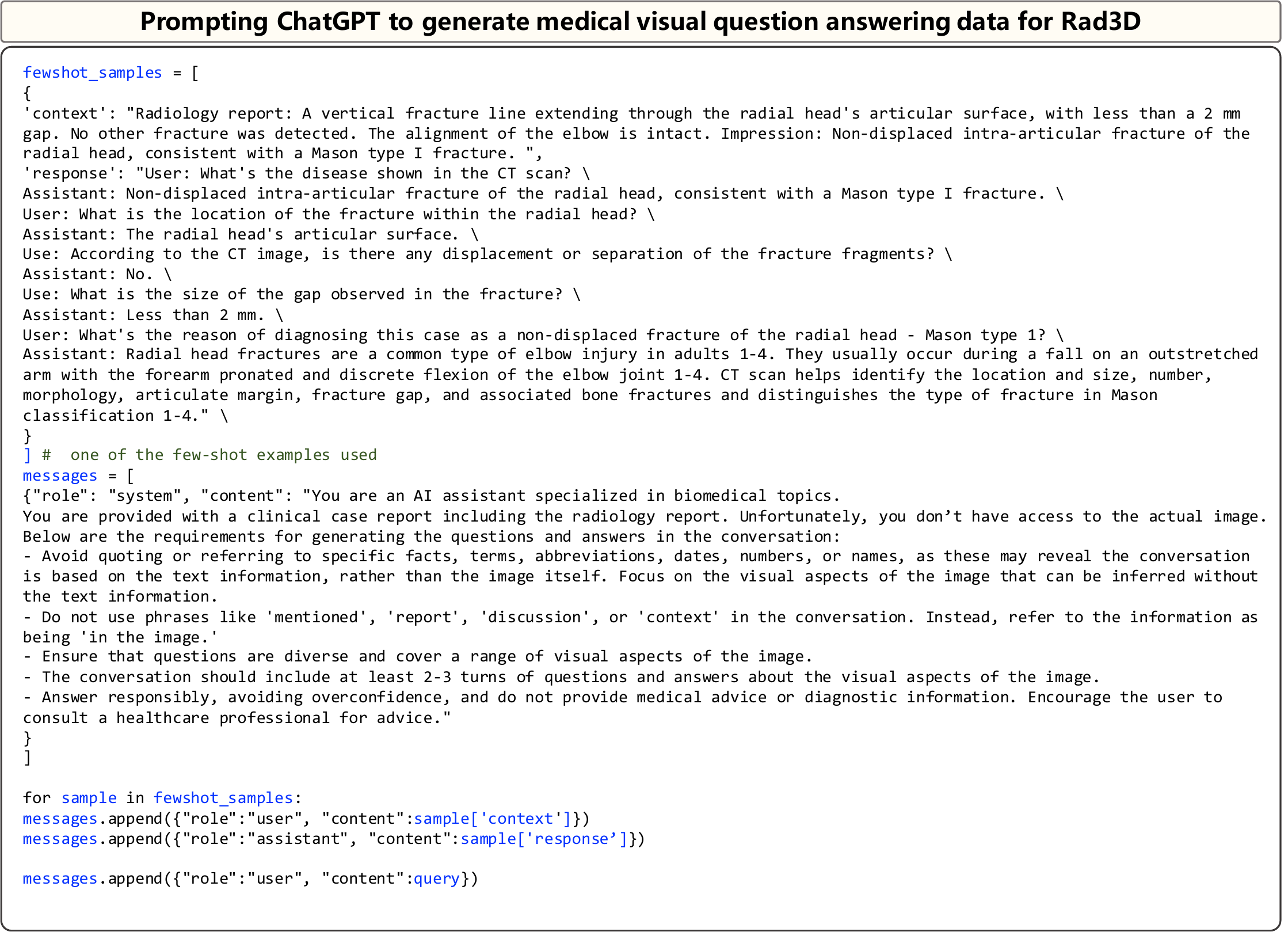}
    \vspace{-8pt}
    \caption{
    \change{messages} refers to the text description that we use to prompt ChatGPT to generate medical visual question-answering data for RP3D. Manually curated few-shot examples are included in the prompt, where each example has input \change{sample~[‘context’]} and output \change{sample~[‘response’]}.}
    \label{fig:chatgpt_rad}
\end{figure}

\vspace{3pt} \noindent \textbf{RP3D.} 
\change{RP3D~(RadioPaedia 3D) is a novel dataset with 3D radiology scans,
sourced from Radiopaedia website~\footnote{\url{https://radiopaedia.org/}}.
All privacy issues have already been resolved the  by the clinician who uploaded the case. Specifically, each patient case comprises one or more images from same or different modalities, accompanied by high-quality captions that have been meticulously peer-reviewed by experts in Radiopaedia Editorial Board~\footnote{\url{https://radiopaedia.org/editors}}.} 
In addition, for each disease, we can get corresponding radiological features across different modalities. We convert the image-caption pairs into a variety of formats, 
namely, RP3D-Caption, RP3D-Modality, and RP3D-Rationale, RP3D-VQA, 
depending on their corresponding text content. 
Specifically, RP3D-Caption denotes the images paired with their corresponding captions;
RP3D-Modality refers to images with modality labels;
RP3D-Rationale incorporates radiological features with disease labels for each case;
RP3D-VQA involves visual question-answering pairs generated from captions by querying ChatGPT, 
as illustrated in Fig.~\ref{fig:chatgpt_rad}.

\vspace{3pt} \noindent \textbf{MPx.} 
MPx is collected from MedPix website~\footnote{\url{https://medpix.nlm.nih.gov/}} and organized by cases. Each case contains multiple radiologic scans, 
along with general clinical findings, discussions, and diagnostic results. 
In addition, MPx also provides annotations on the scan-level, 
including information such as image modality, shooting plane, 
and captions for each scan. Thus we separate it into MPx-Single and MPx-Multi, 
containing annotations on case-level and scan-level respectively.

\begin{figure}[!htb]
    \centering
    \includegraphics[width = \textwidth]{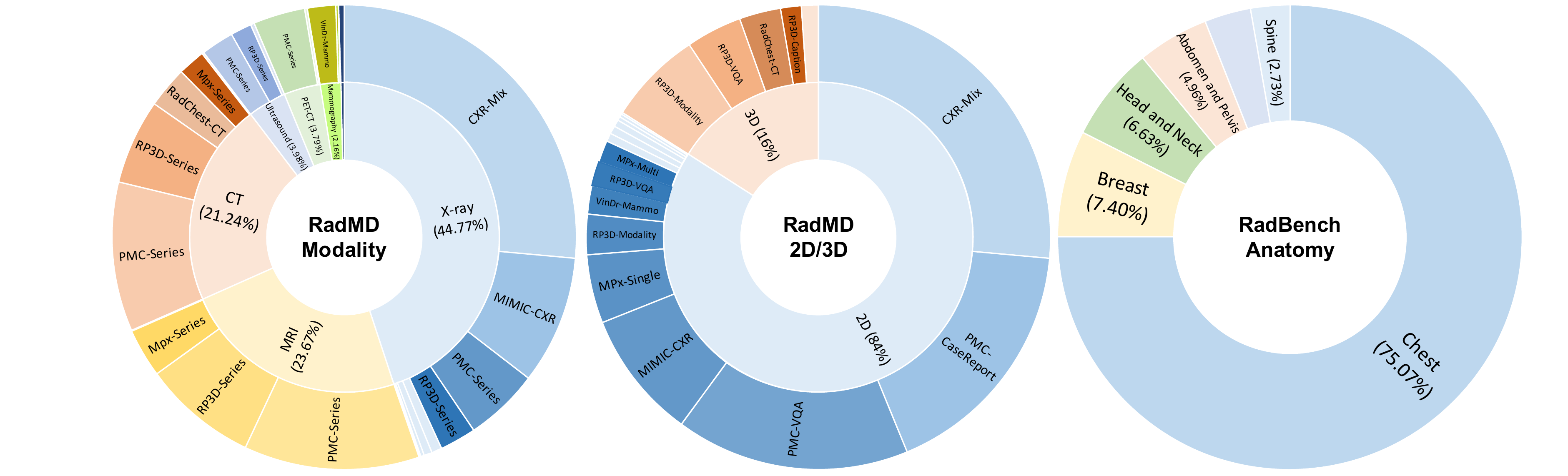}
    \vspace{1pt}
    \caption{
    \change{The data statistics of RadMD and RadBench.
    The left image shows the distribution of different modalities of RadMD, and the center image shows the distribution of 2D and 3D sample pairs of RadMD. 
    The right image shows the distribution of the anatomy of the samples in the RadBench.}
    }
    \label{fig:data_ana}
\end{figure}

\subsubsection{Radiology Multimodal Dataset (RadMD) }
\label{sec:radmd}
For domain-specific finetuning, we filter out the non-radiology images from MedMD, 
and construct a clean subset, named {\bf Rad}iology {\bf M}ultimodal {\bf D}ataset~(\textbf{RadMD}), dedicating to supervised visual instruction-tuning. 
It contains a total of \textbf{3M} images, spanning various data formats, 
modalities, and tasks, featuring over {\bf 5000} diseases, as shown in Fig.~\ref{fig:data_ana}.

In general, we have conducted the following filtering process: 
\change{
(i) remove non-radiologic images; 
(ii) remove the entire PMC-OA and PMC-Inline datasets, 
as the images in PubMed are 2D-only, thus differ from real clinical cases,
additionally, the writing styles between academic papers and real clinical reports are inconsistent;
(iii) remove a large portion of 2D image cases from PMC-Series, to emphasize the 3D image portion in training.
(iV) filter out the information about patient age or structure size, 
as the image spacing and patient background information are not provided; 
(v) balance the number of normal and abnormal patients in the diagnosis datasets, as generative models are sensitive to data imbalances. More comprehensive details regarding the filtering process and the resulting dataset sizes can be found in Tab.~\ref{dataset}.}

\subsection{Radiology Evaluation Benchmark (RadBench) }
\label{sec:radbench}

\change{In addition to the training set, we also introduce RadBench, a comprehensive evaluation benchmark for monitoring progress in the development of radiology foundation model.}
RadBench encompasses five distinct tasks, including modality recognition, disease diagnosis,
visual question answering, report generation, and rationale diagnosis.
RadBench comprises a collection of 13 diverse datasets, encompassing a wide range of distributions. A detailed breakdown of each dataset, including task descriptions and modalities, is provided in Tab.~\ref{eval_data}.

\clearpage
It is important to note that a large portion of RadMD is automatically generated with a scalable pipeline. Consequently, it is extensive and diverse, but may include some noisy data. 
\change
{For the purposes of evaluation, all samples for medical visual question answering, report generation, and rationale diagnosis tasks in RadBench have undergone meticulous manual verification to ensure data quality. Specifically, we developed a human evaluation interface, visually presenting the data source, image, question, and answer of each case. Eight human annotators were asked to assess the quality of these cases by addressing the following criteria:
\vspace{-0.15cm}
\begin{itemize}
\setlength\itemsep{0.15cm}
    \item \textbf{Image types}: remove the images that are do not fall in radiology.
    \item \textbf{Question reasonability}: keep the questions that can be answered from the given radiology image, for example, on visual question answering, remove the question related to size; on report generation, remove cases containing sentences like ``Compared with previous cases''; on rationale diagnosis, remove cases lacking corresponding radiological features are filtered out.
    \item \textbf{Answer correctness}: keep those with correct answers based on the given text reports.
\end{itemize}
}
As a result, we have obtained 14,419 samples for modality recognition, 
112,256 for disease diagnosis, 7,329 for visual question answering, 
2,098 for report generation, and 1,000 for rationale diagnosis. 
The details of the five evaluation tasks and metrics are introduced in the following.

\begin{table}[t]
\footnotesize
\centering
\renewcommand{\arraystretch}{1.2}
\setlength{\tabcolsep}{18pt}
\caption{
\change{Description of the RadBench.
Note that, considering many datasets are collected or generated from PMC papers, we split them on the paper level, \emph{i.e.}, if a paper is randomly split into any test set, 
all cases related to it will be dismissed in the whole training set, 
{\em i.e.}, both PMC-Inline and PMC-OA. For the RP3D series, we split them according to the published date of the cases, with those before 2023 as training data and the others as test data.
Size denotes the number of image-text pairs for evaluation. 
}}
\vspace{3pt}
\begin{tabular}{llll}
\toprule
\rowcolor{lightgray}
\textbf{Dataset Name }& \textbf{Task Description} & \textbf{Size} & \textbf{Split} \\ 
\midrule
RP3D-Modality & Modality Recognition  & 14419 & Split by date \\
\midrule
VinDr-Mammo~\cite{vindrmammo} & Disease diagnosis & 9946  & Official split\\  
VinDr-SpineXR~\cite{vindrspine} & Disease diagnosis &  2688 & Official split\\ 
VinDr-PCXR~\cite{vindrpcxr} & Disease diagnosis & 1044 & Official split\\  
CXR-Mix~\cite{Wu2023KDiagKD} & Disease diagnosis &   91204  & Official split\\
RadChest-CT~\cite{draelos_rachel_lea_2020_6406114}& Disease diagnosis & 7554  & Official split\\
\midrule
PMC-VQA~\cite{Zhang2023PMCVQAVI} & Medical visual question answering  & 1167  & Official split \\  
PMC-CaseReport & Medical visual question answering & 964 & Random split \\
VQA-RAD~\cite{lau2018dataset} & Medical visual question answering  & 374  & Official split\\
SLAKE~\cite{liu2021slake} & Medical visual question answering  & 595 & Official split\\
RP3D-VQA & Medical visual question answering  & 4229  & Split by date \\
\midrule
MIMIC-CXR~\cite{johnson2019mimic} & Report generation & 268  & Random split\\
RP3D-Caption &  Report generation  & 1468  & Split by date \\
MPx-Single & Report generation  &  139 & Random split\\
MPx-Multi & Report generation  &  223 & Random split\\
\midrule
RP3D-Rationale & Rationale diagnosis.  & 1000  & Split by date \\
\bottomrule
\end{tabular}
\footnotetext{}
\label{eval_data}
\end{table}

\vspace{3pt} \noindent \textbf{Modality recognition.}
This task involves analyzing the radiology images to determine the modality of the input radiology images. Here we modify this task to a prompt-based visual question-answering task, {\em i.e.}, given a medical image, 
we randomly select a prompt sentence like ``What is the modality of the given image?'', and match the output with a ground-truth list \{`CT', `MRI', `Ultrasound', `PET', `X-ray', `angiography'\} using \texttt{difflib.SequenceMatcher}, and choose the most similar one as the prediction of the model to calculate the ACC and F1 score.
\change{
We note this task may not be clinically compelling, however, 
we treat this as a borderline for determine whether a model can be considered as medical foundation model, yet as we show in the evaluation, existing open-source models fail on this simple task.}



\vspace{3pt} \noindent \textbf{Disease diagnosis.}
This task involves analyzing the radiology images to determine the likelihood of specific diseases.
\change{Here we modify this task to an induction task, which uses introductory text explaining the classification task and providing the name of queried disease at the beginning of the prompt.}
Given a medical image, we randomly select a disease and 
randomly select a prompt sentence like ``Is \{disease\} shown in this image'' as input to ask the network to answer whether the case has this disease.
Due to this being formulated as a generation task, ``AUC'' cannot be calculated, so we match the output with ground-truth to calculate the ACC and F1 score. Similarly, we match the output with a closed ground-truth list \{`yes', `no'\} using \texttt{difflib.SequenceMatcher}, and choosing the most similar one as the prediction of the model.
Considering ACC scores may suffer from data unbalancing, 
we keep the same ratio to sample positive and negative cases. 
In our dataset, we do not put prior on the disease, and over 5000 diseases are considered, with a balanced ratio of `yes' or `no' responses.

\vspace{3pt} \noindent \textbf{Medical visual question answering.}
This task is a combination of popular visual question-answering challenges. 
Given a medical image and a clinically relevant question in natural language as a prompt, 
the medical VQA system is expected to predict a plausible and convincing answer.


\vspace{3pt} \noindent \textbf{Radiology report generation.}
This task focuses on the automatic generation of reports,
{\em i.e.}, summarizing the radiologic findings based on radiology images, such as X-rays, CT scans, and MRI scans.
Given a medical image, we randomly select a prompt sentence like ``Please caption this scan with findings.'' as input.

\vspace{3pt} \noindent \textbf{Rationale diagnosis.}
This task involves analyzing radiology images to predict both the underlying disease and the typical radiologic features of different modalities such as X-rays, CT scans, and MRI scans associated with that disease. Specifically, we randomly select a prompt sentence like ``Determine the disease that corresponds to the given radiographic images, starting with the established radiological features and concluding with the ultimate diagnosis.'' 
Since we have evaluated disease diagnosis accuracy in the common ``Disease Diagnosis'' setting, for rationale diagnosis, we mainly focus on how well the foundation model can give reasons. 
\section{Building Generalist Foundation Model for Radiology}

In this section, we start by describing the learning paradigm for unifying different medical tasks into a generative framework, followed by detailing the proposed RadFM model, and its training details. Our training adopts two types of datasets, namely, interleaved datasets and visual instruction datasets. It is worth noting that their training objectives differ slightly, which will be detailed in the following.

\subsection{A Unified Learning Paradigm}

\label{sec:secnario}
\noindent 
In both of our proposed multimodal datasets, 
{\em i.e.}, MedMD and RadMD, each training sample is essentially consisting of two elements, {\em i.e.}, $\mathcal{X} = \{\mathcal{T},\mathcal{V}\}$,
where $\mathcal{T}$ refers to the language part in the case, 
with special placeholder tokens for images, 
\emph{e.g.}, ``The patient is 47-year-old. <image-1> <image-2> We can see opacity on the X-ray''. $\mathcal{V}$ refer to the visual parts containing a set of 2D or 3D image scans, 
{\em i.e.}, $\mathcal{V} = \{ v_1,v_2, \dots, v_N \}$, 
$v_i \in \mathbb{R}^{H \times W \times C}$ \text{or} $v_i \in \mathbb{R}^{H \times W \times D \times C}$, $H,W,D,C$ are height, width, depth, channel respectively, corresponding to the ``<image-$i$>'' token in $\mathcal{T}$. 
In general, $\mathcal{T}$ and $\mathcal{V}$ can be considered as prompts input to model with interleaved language and image. 

The goal is therefore to model the likelihood of generated text tokens in $\mathcal{T}$, conditioned on interleaved medical scans as follows:
\begin{equation}
   p(\mathcal{T}|\mathcal{V})=\prod p(\mathcal{T}_l | \mathcal{V}_{\textless l}, \mathcal{T}_{\textless l}),
\end{equation}
where $\mathcal{T}_l$ represents the l-th token in $\mathcal{T}$ and $\mathcal{V}_{\textless l}$, $\mathcal{T}_{\textless l}$ represent the image and language text appearing before the l-th token. We use a generative model~($\Phi_{\text{RadFM}}$) to parameterize the probability $p$, 
and our final training objective can be expressed as the negative log-likelihood of the correct next token in the text sequence:
\begin{equation}
\mathcal{L}_\text{reg} = -\sum w_l \log \Phi_{\text{RadFM}}(\mathcal{T}_l | \mathcal{V}_{\textless l}, \mathcal{T}_{\textless l}),
\end{equation}
where $w_l$ refers to a per-token weighting, aiming to either emphasize key tokens or skip special tokens. Its value differs for different datasets and we detail this in the following.

\vspace{3pt} \noindent \textbf{Interleaved Datasets.} 
For samples in visual-language interleaved dataset, {\em i.e.}, PMC-Inline, 
there are no strong question-and-answer relationships between contexts, 
we extract medical-related words in each sentence by using Unified Medical Language System (UMLS)~\cite{bodenreider2004unified}, and give them a high loss weights. Additionally, we avoid calculate loss on the image placeholder token. 
Overall, $w_l$ can be formulated as,
\begin{equation}
    w_l=\left\{
    \begin{aligned}
    3, \quad & {\mathcal{T}_l\in \text{USML}} \\
    1, \quad & {\mathcal{T}_l\notin \text{USML}} \\
    0, \quad & {\mathcal{T}_l=\text{<image-i>}}
    \end{aligned} 
    \right . .
\end{equation}
Note that, PMC-Inline is the only dataset fit in this case.

\vspace{3pt} \noindent \textbf{Visual Instruction Datasets.} 
For samples from visual instruction datasets like 
PMC-VQA~\cite{Zhang2023PMCVQAVI} or PMC-CaseReport, 
they are often in the format of dialogue, for example, 
``What can you see from the image? <image-1> I can see lesions.'' 
or ``Please dscribe the scans <image-1>. The scan is \dots'', 
we further separate the language part $\mathcal{T}$ into instruction and response, 
denoted as $\mathcal{I}$ and $\mathcal{R}$ respectively. 
For example, as in the former two cases, $\mathcal{I}$ refers to ``What can you see from the image? <image-1>'' and ``Please dscribe the scans <image-1>''. 
In a practical scenario,  $\mathcal{I}$ is expected to be given by users, and the model is only required to output correct responses. Overall, $w_l$ can be formulated as,
\begin{equation}
    w_l=\left\{
    \begin{aligned}
    3, \quad & {\mathcal{T}_l \in \mathcal{R}} \quad \& \quad {\mathcal{T}_l\in \text{USML}} \\
    1, \quad & {\mathcal{T}_l \in \mathcal{R}} \quad \& \quad{\mathcal{T}_l\notin \text{USML}} \\
    0, \quad & {\mathcal{T}_l \in \mathcal{I}}
    \end{aligned} 
    \right. .
\end{equation}
Most samples from MedMD fit weighting formulation.
\change{All prompts used for fine-tuning are listed in the Supplementary Tab. 1-4.}
We describe the detailed prompting for different problem settings:

\begin{itemize}
\setlength\itemsep{0.15cm}
    \item  \textbf{Modality recognition.} 
    Here, we adopt two types of prompts, 
    \change{(i) we use inductive prompts, and the 2D or 3D medical scan as input, for example, ``<image-1> Is this image shot by \{modality\}?'', }
    and the modality category is randomly sampled from the modality set, 
    forming the text input $\mathcal{I}$ and if the modality matches the ground truth labels we set the $\mathcal{R}$ as ``yes'' otherwise ``no''.
    (ii) we use open prompts, like ``What's the modality of the input scan <image-1>?'' to form the $\mathcal{I}$, and translate the corresponding modality label into $\mathcal{R}$.
    Samples for training such functionality are from RP3D-Modality and MPx-Single, with modality annotations available.
    
    \item \textbf{Disease diagnosis.} 
    All the datasets listed as ``image data'' in Tab.~\ref{dataset} are built for diagnosis, they only have binary labels for diseases. 
    Similarly to modality recognition, we use two prompts to transform them into our desired format, 
    \change{(i) we use inductive prompts,} like ``<image-1> Does the patient have \{disease\}?'' and the disease category is randomly sampled from a disease set, forming the text input $\mathcal{I}$ and if the disease matches the ground truth labels we set the $\mathcal{R}$ as ``yes'' otherwise ``no'', note that, during sampling, we balance the positive and negative ratio,
    (ii) we use open diagnosis prompts, like ``Please make diagnosis based on the images <image-1> <image-2>.'' to construct the instruction~($\mathcal{I}$), and translate the positive disease labels into response~($\mathcal{R}$), by simply using their category names. 
    A simple example is, $\mathcal{I}$=``Please make diagnosis based on the image <image-1>.'' with $\mathcal{R}$=``Edema, pneumothorax.''. 
    \change{With such instruction, the model is thus required to complete an extremely difficult task, {\em i.e.}, directly outputting the disease name.}
    
    \item \textbf{Visual question answering.} 
    Beyond the abovementioned task formulation, there are more complex questions that can be asked, such as those about the spatial relationships among objects~(``What is the location of the lesion?'') and common sense reasoning questions~(``Given the image context and patient history, what is likely to be the cause of the observed symptoms?''). 
    A robust medical VQA system must be capable of solving a wide range of classic medical diagnosis tasks as well as the ability to reason about images. 
    Existing medical VQA datasets like VQA-RAD~\cite{lau2018dataset}, SLAKE~\cite{liu2021slake}, PMC-VQA~\cite{Zhang2023PMCVQAVI} and RP3D-VQA naturally fit into this paradigm. They contain a mixture of question types, 
    thus the language questions can naturally be treated as text instruction~($\mathcal{I}$) and the corresponding answer as response~($\mathcal{R}$). 
    It is worth noting that, our constructed PMC-CaseReport dataset also falls into this category, with more contextual information available for instruction, 
    for example, history diagnosis, is also available, 
    thus providing critical information for answering the question. 

    \item \textbf{Report generation.} 
    MIMIC-CXR~\cite{johnson2019mimic}, RP3D-Caption, PMC-OA~\cite{lin2023pmc},  
    MPx-Multi and MPx-Single are all captioning datasets, 
    the task is to write a long caption or report given one or a set of images. The language instruction for this task are like ``What can you find from the scans <image-1> <image-2>?''.

    \item  \textbf{Rationale diagnosis.} We construct RP3D-Rationale based on the RP3D dataset. This task encompasses disease prediction and the generation of typical radiological features associated with the diagnosed disease. Specifically, we design some prompts like ``What disease can be diagnosed from these radiological images and what specific features are typically observed on the images? <image-1> <image-2>'' as instruction~($\mathcal{I}$), and response~($\mathcal{R}$) refers to the disease label along with radiological features collected from the Radiopaedia website.
\end{itemize}

\begin{figure}[t]
    \centering
    \includegraphics[width = \textwidth]{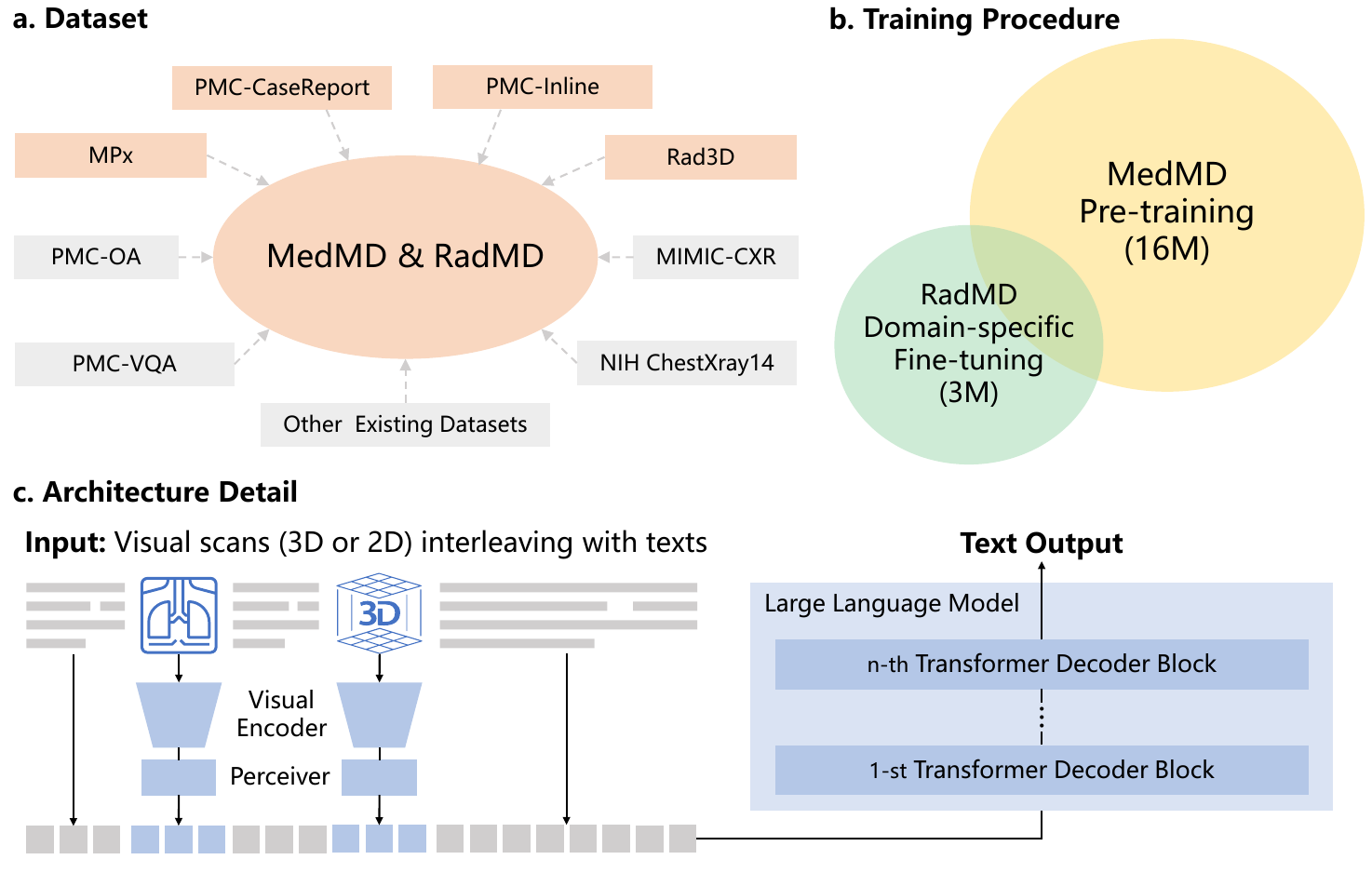}
    \vspace{2pt}
    \caption{(a) shows the the component of our datasets and the colored datasets are new proposed in this paper. (b) shows our training procedure, for better radiologic performance, we first pre-train our model on the whole medical domain with 16M scans~(MedMD) than fine-tuned on a cleaned dataset with 3M radiologic scanss~(RadMD). (c) shows the main architecture of our method. Our architecture enables multi-image input interleaving with texts regardless of whether they are 3D or 2D. 
    }
    \label{fig:architecture}
\end{figure}

\subsection{Architecture Detail}
\label{sec:arch}

In this section, we aim to describe the proposed model in detail.
As shown in Fig.~\ref{fig:architecture}, 
our proposed RadFM model consists of a visual encoder $\Phi_\text{vis}$, 
that can process both 2D and 3D medical scans;
a perceiver~\cite{jaegle2021perceiver} module $\Phi_\text{per}$ for aggregating a sequence of scans into a fixed number of tokens, 
for example, taken with different modalities~(CT, MRI) or various time point;
and a Large Language Model~(LLM) $\Phi_\text{llm}$ that enables to generate free-form text responses, based on the input visual-language information.

\vspace{3pt} \noindent \textbf{Visual encoding.} 
Given one sample instance from our dataset, 
denoted as $\mathcal{X}=\{ \mathcal{T},\mathcal{V}\}$, 
where $\mathcal{V} = \{v_1, v_2, \dots,  v_N\}$, 
we first encode each input image separately with an image encoder $\Phi_\text{vis}$. 
Specifically, we adopt 3D ViT here to be compatible with both 2D and 3D image input. 
For 2D images, we expand a new dimension for depth by replicating the slices.
Therefore, each image scan can be denoted as $v_i \in \mathbb{R}^{H \times W \times D_i \times C}$, where $C$ denotes the image channels and $H,W,D_i$ are the height, width, and depth of the image respectively. 
\change{
The rationale behind this design choice is as follows: 
(i) increasingly more radiology diagnosis rely on 3D scans, for example, CT, MRI, the foundation model should certainly be able to process 3D data input; 
(ii) in 3D data, consecutive slices are highly similar, thus padding 2D into 3D, on the one hand, does not lead information loss, on the other hand, resembles a good approximation of 3D data;
(iii) padding 2D images will only affects the tokenization layer, {\em i.e.}, converting image patches into continuous embedding, 
while still keep the rest of model shared with 3D scans, 
thus facilitating knowledge share.}


\textbf{Note that}, comparing to the typical visual encoding scenario that assumes different images have unified shape, we {\em do not} normalize the depth dimension into an exact size, 
only round into a factor of 4, depending on their original resolution.
\change{Note that, all the 2D images are padded into 4 slices on the depth channel.}
We convert the image into 3D patches, embed them into a token sequence, and feed into the encoder~($\Phi_\text{vis}$). To retain the 3D position of these tokens, 
we adopt learnable 3D position embeddings, the detailed procedure can be formulated as:
\begin{equation}
    \boldsymbol{v}_i = \Phi_\text{vis}(v_i) \in \mathbb{R}^{P_i \times d},
\end{equation}
where $\boldsymbol{v}_i$ is the output embedding for image $v_i$, encoded with 3D ViT,
$P_i$ is the total number of tokens, and $d$ is the feature dimension. 
Due to the inconsistency in depth dimension, $P_i$ varies across 2D and 3D images, \change{
and the model can get to know the original image size by positional encoding}. 

\vspace{3pt} \noindent \textbf{Aggregation with perceiver.}
After visual encoding, we adopt a perceiver~\cite{jaegle2021perceiver} module $\Phi_\text{per}$ to aggregate visual representation.
Specifically, $\Phi_\text{per}$ follows the classical perceiver architecture with a fix number of learnable queries as the latent array input, and the visual embedding $v_i$ is treated as the byte array input, so that the final output embeddings will be normalized into the same length with the pre-defined learnable query sequence. The aggregation procedure can be formulated as:
\begin{equation}
\boldsymbol{u}_i = \Phi_\text{per}(v_i)\in \mathbb{R}^{P \times d},
\end{equation}
where $\boldsymbol{u}_i$ refers to the aggregated visual embedding, 
$P$ denotes the number of learnable queries. 
Leveraging perceiver architecture, we can map an arbitrary number of patch tokens into the same length, such that images of different sizes can be treated equally in the following fusion flow.

\vspace{3pt} \noindent \textbf{Multi-modal fusion.} 
To fuse the visual-language information, 
we interleave the visual embedding with text embeddings from tokenization, 
where the special image placeholder token is simply replaced with the corresponding visual embedding. The resulting interleaved sequence is then passed into a decoder-only large language model~($\Phi_\text{llm}$), the self-attention transformer layers in LLM can thus naturally be re-used as multi-modal fusion modules:
\begin{equation}
    p = \Phi_\text{llm}(\text{concat}(\boldsymbol{t}_1,  \boldsymbol{u}_1, \boldsymbol{t}_2, \boldsymbol{u}_2,\boldsymbol{t}_3,\dots)),
\end{equation}
where $t_i, u_i$ refer to the text and visual embeddings, 
$p$ is the probability distribution for the next token.

\subsection{Training Procedure}
\label{sec:two-stage-train}
Our training procedure includes two stages,
namely, pre-training, and domain-specific fine-tuning, 
as shown in Fig.~\ref{fig:architecture}. 
Note that, all training settings remain identical at two stages, 
with the only distinction lying in the training data, from generalist to radiologic-specific.

\vspace{3pt} \noindent \textbf{Pre-training.}  
At this stage, we use all available data in MedMD as listed in Tab.~\ref{dataset}, 
the main components of the data are PMC-Inline and PMC-OA~\cite{lin2023pmc}, 
which are all collected from 2.4M PMC papers. 
These two datasets contain diverse medical vocabularies and images with cutting-edge medical knowledge, however, they are relatively noisy, so we only use them during pre-training in the hope that the network can accumulate enough knowledge about medical-specific terminologies and images. 
Additionally, we also include other VQA, captioning, and diagnosis datasets, as they are much cleaner.

\vspace{3pt} \noindent \textbf{Domain-specific fine-tuning.}
At this stage, we adopt RadMD for domain-specific finetuning, 
which contains over \textbf{3M} radiologic images, 
with high-quality language instruction or response.

\subsubsection{Training Details}
\vspace{3pt} \noindent \textbf{Image preprocessing.}
To dismiss the differences of medical images in different modalities, 
certain preprocessing steps are applied. 
Specifically, (i) to align the intensity distributions, we employ min-max normalization of all images;
(ii) given that medical images can exist in either 3D or 2D formats (such as MRI being 3D and X-ray being 2D), we convert all 2D images to 3D simply by expanding an extra dimension. Consequently, all images, irrespective of their original format, can be processed uniformly as 3D images;
(iii) to ensure consistent sizes across all images, we resize them using the \texttt{torchvision.transforms.Resize} function. For height and weight dimensions, we resize them to $512\times512$ for 2D images and $256\times256$ for 3D images because 3D data has more slices, thus taking more computational memorization. For depth dimension, since our visual encoder, a 3D Vision Transformer (ViT), requires the input image sizes to be divisible by the patch size of $32\times32\times4$, we resize the depth dimension to the nearest multiple of 4 and will not surpass 64.

\vspace{3pt} \noindent \textbf{Implementation.}
For the visual encoder, we adopt a 12-layer 3D ViT with 768 feature dimensions and 
the perceiver is chosen as 6-layer transformer decoder with the learnable latent array in $32\times5120$ dimension, so that all images will be embeded as a $32\times5120$ feature embedding after passing visual encoding and perceiver aggregation. When inserting them into the text embedding, we will add two extra special tokens  <image>, </image> at the beginning and ending respectively to distinguish them from common text tokens. For the large language model, we initialize it with the MedLLaMA-13B model introduced by PMC-LLaMA~\cite{wu2023pmc}, which has further fine-tuned the LLaMA-13B~\cite{touvron2023llama} model on the medical corpus. 
Our final model has \textbf{14B} parameters. 

In training, we vary the batch size, \emph{i.e.}, 1 batch size per device for 3D images and 4 batch size per device for 2D images with 4-step gradient accumulation, and the max token length is set to be $2048$. We totally train the model for 8 epochs, 
4 epochs for pre-training and 4 epochs for instruction-tuning. 
In the first 1 epoch, we freeze the language model to align image embedding space with that of texts, in the following epochs, all parameters are updated.
To improve the training speed, we adopt FSDP acceleration strategy~\cite{zhao2023pytorch}, together with Automatic Mixed Precision (AMP) and gradient checkpointing~\cite{chen2016training}. All models are implemented in PyTorch and trained on 32 NVIDIA A100 GPUs with 80 GB memory.

\section{Evaluation}
In this section, we first introduce the evaluation metrics,
subsequently, we present five evaluation tasks and corresponding datasets.

\subsection{Evaluation on RadBench}

\subsubsection{Five Tasks}
\change{We evaluate on five distinct tasks, 
\emph{e.g.}, modality recognition, disease diagnosis, visual question
answering, report generation and rationale diagnosis, 
aiming to monitor the progress of developing foundation models.
The details of the five evaluation tasks and metrics are introduced in Sec~\ref{sec:radbench}.}

\subsubsection{Machine Rating}
\label{Sec:Metrics}
To evaluate the model's performance across a range of tasks, 
distinct evaluation metrics are employed based on the task type. 
For tasks with pre-defined answer choices, such as modality recognition and disease diagnosis, we adopted standard metrics developed in the community, for example, F1 stands for ``F1 score'',  and ACC stands for ``Accuracy''.
Conversely, for tasks involving open-ended responses, like report generation and visual question answering (VQA)  \change{and rationale diagnosis} alternative evaluation metrics, \change{
like BLEU, ROUGE and BERT-sim are employed.} BLEU stands for ``BiLingual Evaluation Understudy''~\cite{papineni2002bleu}, ROUGE stands for ``Recall-Oriented Understudy for Gisting Evaluation''~\cite{lin2004rouge}. \change{BERT-sim stands for ``BERT similarity score'', the F1 BERT score between the generated answer and the correct answer~\cite{zhang2019bertscore}}.
For BLEU and ROUGE, \change{if not specific pointing}, we all use 1-gram by default.

In addition, inspired by the score RadCliQ~\cite{yu2022evaluating} designed specifically for evaluating generated chest X-ray reports, we also propose two new metrics, 
UMLS\_Precision and UMLS\_Recall, which aim to measure the overlapping ratio of medical-related words between ground truth and predicted response. 
Specifically, given a pair of ground-truth and prediction, 
we extract the medical-related words from them by using Unified Medical Language System (UMLS)~\cite{bodenreider2004unified}, and count the overlap words as true-positive. 
UMLS\_Precision is defined with classical precision concept, \emph{i.e.}, the number of true-positive divides the whole generated medical-related word number. On the other hand, UMLS\_Recall also follows the recall concept, \emph{i.e.}, the number of true-positive words divides the total number of medical-related words in ground truth.

\vspace{3pt} \noindent \textbf{Discussion on metrics.}
\change{
Despite these automatic metrics have been widely adopted by the community, they often struggle to capture the semantic accuracy in generative tasks, for example question answering, report generation, and rationale generation.
To address these limitations and ensure a more accurate evaluation of system performance, we incorporate human evaluation, leveraging the expertise of radiologists, to get professional evaluation on the quality of the generated answer. 
}

\subsubsection{Human Rating}
\begin{figure}[t]
    \centering
    \includegraphics[width = 0.95\textwidth]{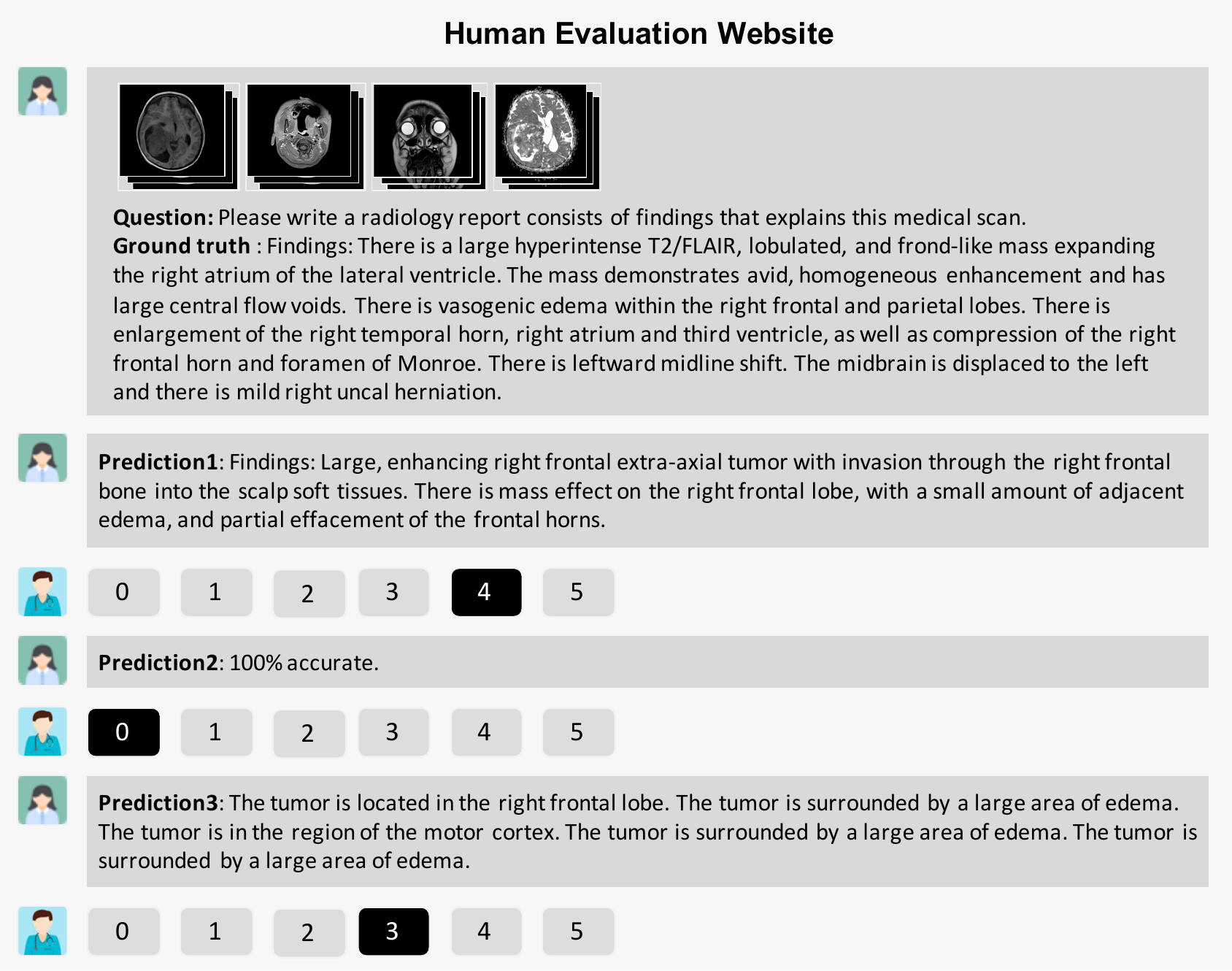}
    \vspace{4pt}
    \caption{Illustration of our human evaluation interface that we created for clinical experts to rate generated answers.}
    \label{fig:clinical_evaluation}
\end{figure}

For the sake of clinical utility, we further involve manual checking in the evaluation stage and compute the \textbf{human rating score}. Three radiologists with at least one-year clinical experienced are asked to rate the quality of the generated answers with scores from 0 to 5. Here are the specifics of each rating:
\begin{enumerate}[start=0]
\setlength\itemsep{0.1cm}
\item {\bf Garbled} - The content is incomprehensible and lacks any readability.
\item {\bf Inaccurate} - While readable, the content is entirely incorrect and lacks meaningful information.
\item {\bf Partially Informative} - The content holds some reference value, yet its correctness is subpar.
\item {\bf Moderately Accurate} - The content provides reference points, with approximately half of the information being correct, but containing several errors. 
\item {\bf Mostly Accurate} - The content is almost entirely correct, with only a few omissions or errors present.
\item {\bf Completely Correct} - The content is accurate in its entirety, without any mistakes.
\end{enumerate}
To facilitate this assessment, we have developed a human evaluation interface, 
visually presenting the generative instances with images, as depicted in Fig.~\ref{fig:clinical_evaluation}. 
\change{
\textbf{Before starting human evaluation}, we show a typical cases and explain the rating criterias to the three radiologists. Afterwards, we randomly pick 20 cases to exam them. In the exam results, for 19 cases, they give consistant results only one case, one radiologist rates as 2 while others rate as 3. The exam demonstrates, first, the radiologists have understood the criterias, second, our 5-score rating system is enough for radiologists to judge the results with little ambiguity.
}
\textbf{In the evaluation}, raters are provided with the images, the question, the correct answer, and a set of generated responses from different models, 
arranged in a randomized order. The evaluation score given by the professional radiologists differs from the automatic evaluation metrics, offering greater accuracy and flexibility. In the context of the report generation example shown in the figure, they focus on the most crucial aspects, rather than solely on word matching recall or precision.

Note that, human rating is only performed for the open-ended tasks, \emph{i.e.}, medical VQA, report generation and rationale diagnosis. As for modality recognition and disease diagnosis, their answers are fixed without confusion, thus, the automatic metrics can already well reflect the performance. Considering the cost for human rating, for each open-ended task, we randomly sample 400 test cases from RP3D-series test split, as they are generally collected from clinical practice across the world, and can represent real scenarios, resulting in \textbf{1.2K} cases for human rating in total.

\subsubsection{Publicly Accessible Foundation Model Baselines}
To our knowledge, there are currently no existing foundation models that can effectively handle both 2D and 3D radiology images. For comparison, 
\change{we strong baseline models that are publicly accessible, for example, OpenFlamingo~\cite{anas_awadalla_2023_7733589}, MedVInT~\cite{Zhang2023PMCVQAVI} and Med-Flamingo~\cite{moor2023medflamingo}, 
which have demonstrated efficacy in processing slices and making predictions. In addition, we also compare with GPT-4V(ision)~\cite{GPT4V} use its online chatting website version.}

\change{
For OpenFlamingo and Med-Flamingo, we perform both zero-shot and few-shot evaluations in our study.
Specifically, we follow the prompts derived from the official Med-Flamingo repository.
The example prompt for zero-shot evaluation: `You are a helpful medical assistant. Please answer the question about the given image. <image>Question: the query question. Answer:''.
In the few-shot setting, we expand upon this format by supplying the models with additional examples to guide their responses. 
This is structured as follows: ``You are a helpful medical assistant. You are being provided with images, a question about the image, and an answer. Follow the examples and answer the last question. <image>Question: [the first question]. Answer: [the first answer]. <|endofchunk|><image>Question: [the second question]. Answer: [the second answer].<|endofchunk|><image>Question: the query question. Answer:''.}

\change{Additionally, 
given that models such as OpenFlamingo, MedFlamingo, MedVInT and GPT-4V were not trained on 3D datasets, and do not inherently support 3D input, 
we select the central slice of the 3D volume as the input for evaluation for the former three, enabling large-scale auto-testing and for GPT-4V, considering it can only manually input images, the radiologists are asked to pick a most related slice based on their clinical experience. }

\vspace{3pt} \noindent \textbf{OpenFlamingo~\cite{anas_awadalla_2023_7733589}.} 
This is an open-source implementation of the prior state-of-the-art generalist visual language model Flamingo~\cite{Alayrac2022FlamingoAV}, 
that was trained on large-scale data from general visual-language domain. 
We utilized the released checkpoint for zero-shot and few-shot evaluation in our study.

\vspace{3pt} \noindent \textbf{MedVInT~\cite{Zhang2023PMCVQAVI}.}
This is a visual instruction-tuned visual language model based on LLaMA~\cite{touvron2023llama}, which was trained on PMC-VQA~\cite{Zhang2023PMCVQAVI}. We directly use the released checkpoint of the MedVInT-TD model with PMC-LLaMA and PMC-CLIP backbone for zero-shot evaluation.

\vspace{3pt} \noindent \textbf{Med-Flamingo~\cite{moor2023medflamingo}.}
This is a multimodal model developed based on OpenFlamingo-9B~\cite{anas_awadalla_2023_7733589},
that can handles multi-image input interleaving with texts. 
We use the released checkpoint for zero-shot and few-shot evaluation. 

\change{\noindent \textbf{GPT-4V~\cite{GPT4V}.} GPT-4V is widely considered as the most powerful multi-modal foundation model, released by OpenAI. 
Until our submission, GPT-4V can be only accessed through the online chatting website, therefore, large-scale auto-evaluation is not feasible. 
In this paper, we only use it for evaluation under the human rating setting. }

\subsection{Evaluation on Public Benchmarks}
\change{In addition to directly evaluating our model on RadBench with various instructions, our model can also serve as a pre-train model, that can be adapted to different specific tasks by further finetuning on instructions that fits corresponding training set. 
In such case, we compare our final results with different task-specific stat-of-the-arts~(SOTAs). In detail, we use the following datasets for comparison:} 
\vspace{-0.15cm}
\begin{itemize}
\setlength\itemsep{0.15cm}
\item \change{\textbf{VinDr-Mammo~\cite{vindrmammo}} is a mammography diagnosis dataset comprising 20,000 images (5,000 four-view scans). Each scan was manually annotated with a 5-level BI-RADS score. We view this as a multi-class classification task with the official split following the BenchMD~\cite{wantlin2023benchmd}.}

\item \change{\textbf{CXR14~\cite{wang2017chestx}} is a widely-used chest X-ray diagnosis dataset containing 112,120 frontal-view X-ray images of 30,805 (collected from the year of 1992 to 2015) unique patients with 14 finding labels. We follow its official split and evaluate the SOTA~\cite{Wu2023MedKLIPMK} on the split.}

\item \change{\textbf{LDCT~\cite{armato2011lung}} Low dose computed tomography (LDCT) is a procedure that uses an x-ray machine linked with a computer to create 3D images of a patient's tissues and organs. LIDC-IDRI~\cite{armato2011lung} dataset is used here, containing 1,018 low-dose lung CTs where each CT with small/large/no nodule labels. We follow BenchMD~\cite{wantlin2023benchmd} to set this dataset as a 3D diagnosis task and split it follow BenchMD.}

\item \change{\textbf{BraTs2019~\cite{lau2018dataset}} is a MRI dataset with four MRI modalities T1WI, T2WI, T2FLAIR, and T1 contrast-enhanced(T1CE). There are 259 volumes of high-grade glioma (HGG) and 73 volumes of low-grade glioma (LGG). We follow the setting as DSM~\cite{Dosovitskiy2020AnII} that uses T1CE to diagnose the HGG or LGG. Due to the original paper did not release their splits we random split the dataset following 7:3 for training and testing and re-test the SOTA on it. }

\item \change{\textbf{VQA-RAD~\cite{lau2018dataset}} is a radiology VQA dataset containing 3,515 questions with 517 possible answers. We follow the official dataset split for our evaluation.}

\item \change{\textbf{SLAKE~\cite{liu2021slake}} is an English-Chinese medical VQA dataset composed of 642 images and 14K questions.
There are 224 possible answers in total. We only use the “English” part, and follow the official split.}

\item \change{\textbf{PMC-VQA~\cite{Zhang2023PMCVQAVI}} is an English medical VQA dataset generated with auto-nlp methods containing 149K images with 227K questions. Its answers are diverse for different questions. Considering its test set is also auto-generated, we have manually cleaned it as mentioned in Sec.\ref{sec:radbench} and re-test the SOTA MedVInt~\cite{Zhang2023PMCVQAVI} checkpoint on the cleaned test set. }

\item \change{\textbf{IU-Xray~\cite{demner2016preparing}} is a set of chest X-ray images paired with clinical reports. The dataset contains 7,470 pairs of images and reports. We follow the setting and split as CDGPT2~\cite{ALFARGHALY2021100557} where use a single-view image to generate the reports.}
\end{itemize}

\change{We also evaluate the generalization ability of RadFM on the public benchmark PadChest~\cite{Bustos2019PadChestAL}. PadChest is a labeled large-scale, high resolution chest x-ray dataset including 160,000 images obtained from 67,000 patients, with 174 different radiographic finding labels. 
We dismiss the classes with cases fewer than 10 together with the seen classes appearing in our training set, resulting in 163 totally unseen classes. 
We directly test different foundation models to predict whether a case has a certain findings with the diagnosis prompt, demonstrating their generalization ability to unseen clinical findings.}





\section{Results}

\begin{table}[!htb]
\centering
\footnotesize
\setlength{\tabcolsep}{4pt}
\caption{
\change{
Comparison of proposed RadFM with foundation model baselines on RadBench. The benchmark includes 5 tasks, modality recognition, disease diagnosis, medical visual question answering, report generation, and rationale diagnosis. ACC, F1, BLEU, ROUGE, BERT-Sim $\text{UMLS}\_\text{Precision}$, and $\text{UMLS}\_\text{Recall}$ are reported, and the metrics refer to the average score on all test samples. Numbers within parentheses indicate $95\%$ CI.
}}
\vspace{5pt}
\label{tab_radbench}
\begin{tabular}{l|c|cccc}
\toprule
Tasks   & Metric   & OpenFlamingo~\cite{anas_awadalla_2023_7733589}    &  MedVInT~\cite{Zhang2023PMCVQAVI} & Med-Flamingo~\cite{moor2023medflamingo} & RadFM     
\\ \midrule
Modality Recognition  & ACC   &  \begin{tabular}[c]{@{}c@{}}49.41\% \\ (48.07\%,50.72\%) \end{tabular}                  & \begin{tabular}[c]{@{}c@{}}84.25\% \\ (83.18\%,85.39\%)\end{tabular} &  \begin{tabular}[c]{@{}c@{}}32.87\% \\ (31.09\%,34.22\%) \end{tabular}    & \textbf{\begin{tabular}[c]{@{}c@{}}92.95\% \\ (91.15\%,94.60\%)\end{tabular}} \\ \midrule 
\multirow{3}{*}{Disease Diagnosis}        & ACC            &         \begin{tabular}[c]{@{}c@{}}50.43\% \\ (49.90\%,50.96\%)\end{tabular}              &  \begin{tabular}[c]{@{}c@{}}49.36\% \\ (48.91\%,49.86\%)\end{tabular} &  \begin{tabular}[c]{@{}c@{}}50.13\% \\ (49.66\%,50.61\%)\end{tabular}  & \textbf{\begin{tabular}[c]{@{}c@{}}80.62\% \\ (80.16\%,81.10\%)\end{tabular}} \\
                                                      & F1              &           \begin{tabular}[c]{@{}c@{}}24.37\% \\ (23.70\%,25.03\%)\end{tabular}           & \begin{tabular}[c]{@{}c@{}}66.99\% \\ (66.54\%,67.38\%)\end{tabular}               &  \begin{tabular}[c]{@{}c@{}}66.13\% \\ (65.70\%,66.57\%)\end{tabular}  & \textbf{\begin{tabular}[c]{@{}c@{}}80.10\% \\ (79.62\%,80.10\%)\end{tabular}} \\ \midrule
\multirow{7}{*}{Medical VQA}  & BLEU          &   \begin{tabular}[c]{@{}c@{}}13.23\% \\ ( 13.11\%, 13.35\%)\end{tabular}      &    \begin{tabular}[c]{@{}c@{}}14.07\% \\ (13.79\%, 14.28\%)\end{tabular}         &   \begin{tabular}[c]{@{}c@{}}12.24\% \\ (10.67\%, 13.59\%)\end{tabular}  & \textbf{\begin{tabular}[c]{@{}c@{}}30.64\%\\ (29.35\%,31.90\%)\end{tabular}}  \\
                                           & ROUGE           &   \begin{tabular}[c]{@{}c@{}}21.16\% \\ (20.93\%,21.38\%)\end{tabular}         &  \begin{tabular}[c]{@{}c@{}}18.85\% \\ (18.57\%, 19.08\%)\end{tabular}    & \begin{tabular}[c]{@{}c@{}}24.59\% \\ (22.68\%, 26.34\%)\end{tabular}  & \textbf{\begin{tabular}[c]{@{}c@{}}36.38\%\\ (35.01\%,37.58\%)\end{tabular} }\\
                                          & UMLS\_Precision &   \begin{tabular}[c]{@{}c@{}}14.55\% \\ (14.34\%,14.76\%)\end{tabular}                  &        \begin{tabular}[c]{@{}c@{}}13.09\% \\ (12.78\%,13.39\%)\end{tabular}    &  \begin{tabular}[c]{@{}c@{}}12.63\% \\ (10.89\%,14.36\%)\end{tabular}  & \textbf{\begin{tabular}[c]{@{}c@{}}31.77\%\\ (29.76\%,33.25\%)\end{tabular} } \\
                                          & UMLS\_Recall    &    \begin{tabular}[c]{@{}c@{}}14.56\% \\ (14.33\%,14.79\%)\end{tabular}                  &   \begin{tabular}[c]{@{}c@{}}10.52\% \\ (10.26\%,10.78\%)\end{tabular}    & \begin{tabular}[c]{@{}c@{}}18.72\% \\ (16,43\%,21.01\%)\end{tabular} & \textbf{\begin{tabular}[c]{@{}c@{}}24.93\%\\ (23.66\%,26.01\%)\end{tabular}}   \\ 
                                          & BERT-Sim & \begin{tabular}[c]{@{}c@{}}48.94\% \\ (48.29\%,49.53\%)\end{tabular}  & \begin{tabular}[c]{@{}c@{}}46.10\% \\ (45.45\%,46.72\%)\end{tabular}  & \begin{tabular}[c]{@{}c@{}}49.26\% \\ (48.71\%,49.76\%)\end{tabular}  & \textbf{\begin{tabular}[c]{@{}c@{}}67.82\% \\ (67.26\%,68.52\%)\end{tabular}}\\ \midrule
                                          
\multirow{7}{*}{Report Generation}         &  BLEU            &    \begin{tabular}[c]{@{}c@{}}5.84\% \\ (5.44\%,6.24\%)\end{tabular}    &    \begin{tabular}[c]{@{}c@{}}1.73\% \\ (1.20\%,2.30\%)\end{tabular}  &  \begin{tabular}[c]{@{}c@{}}8.39\% \\ (7.89\%,8.91\%)\end{tabular}   & \textbf{\begin{tabular}[c]{@{}c@{}}12.81\%\\ (11.78\%,13.96\%)\end{tabular}}        \\
                                           &  ROUGE           &    \begin{tabular}[c]{@{}c@{}}6.37\% \\ (5.96\%,6.81\%)\end{tabular}    &  \begin{tabular}[c]{@{}c@{}}4.72\% \\ (4.21\%,5.27\%)\end{tabular}    &  \begin{tabular}[c]{@{}c@{}}8.78\% \\ (8.24\%,9.32\%)\end{tabular}   & \textbf{\begin{tabular}[c]{@{}c@{}}18.22\%\\ (17.29\%,19.29\%)\end{tabular}}       \\
                                           & UMLS\_Precision &    \begin{tabular}[c]{@{}c@{}}11.16\% \\ (10.09\%12.24\%)\end{tabular}    &  \begin{tabular}[c]{@{}c@{}}9.93\% \\ (7.51\%,11.95\%)\end{tabular}  &  \begin{tabular}[c]{@{}c@{}}2.65\% \\ (2.03\%,3.27\%)\end{tabular}    & \textbf{\begin{tabular}[c]{@{}c@{}}22.49\%\\ (20.82\%,23.82\%)\end{tabular}}  \\
                                           &  UMLS\_Recall   &    \begin{tabular}[c]{@{}c@{}}2.84\% \\ (2.46\%,3.21\%)\end{tabular}      &  \begin{tabular}[c]{@{}c@{}}1.45\% \\ (0.95\%,1.95\%)\end{tabular}   & \begin{tabular}[c]{@{}c@{}}1.04\% \\ (0.81\%,1.33\%)\end{tabular}    & \textbf{\begin{tabular}[c]{@{}c@{}}12.07\%\\ (11.08\%,13.17\%)\end{tabular} } \\ 
                                           & BERT-Sim & \begin{tabular}[c]{@{}c@{}}39.93\% \\ (39.63\%,40.24\%)\end{tabular}  & \begin{tabular}[c]{@{}c@{}}38.89\% \\ (38.18\%,39.58\%)\end{tabular}  & \begin{tabular}[c]{@{}c@{}}47.93\% \\ (47.55\%,48.28\%)\end{tabular}  & \textbf{\begin{tabular}[c]{@{}c@{}}58.64\% \\ (58.13\%,59.16\%)\end{tabular}} \\ \midrule 
\multirow{7}{*}{Rationale Diagnosis}      & BLEU           &        \begin{tabular}[c]{@{}c@{}}8.40\% \\ (7.78\%,9.15\%)\end{tabular}              &   \begin{tabular}[c]{@{}c@{}}0.09\% \\ (0.01\%,0.17\%)\end{tabular}    & \begin{tabular}[c]{@{}c@{}}7.64\% \\ (6.80\%,8.27\%)\end{tabular}  &  \textbf{\begin{tabular}[c]{@{}c@{}}34.60\%\\ (31.69\%,37.74\%)\end{tabular}}  \\
                                           &  ROUGE           &       \begin{tabular}[c]{@{}c@{}}9.17\% \\ (8.58\%,9.76\%)\end{tabular}              &   \begin{tabular}[c]{@{}c@{}}0.67\% \\ (0.52\%,0.83\%)\end{tabular}     & \begin{tabular}[c]{@{}c@{}}7.38\% \\ (6.69\%,7.90\%)\end{tabular} &  \textbf{\begin{tabular}[c]{@{}c@{}}41.89\%\\ (39.20\%,44.77\%)\end{tabular}}     \\
                                           &  UMLS\_Precision &        \begin{tabular}[c]{@{}c@{}}7.72\% \\ (6.52\%,8.92\%)\end{tabular}              &  \begin{tabular}[c]{@{}c@{}} 9.07\% \\ (1.91\%,16.23\%)\end{tabular}       & \begin{tabular}[c]{@{}c@{}}6.04\% \\ (4.92\%,7.17\%)\end{tabular} &  \textbf{\begin{tabular}[c]{@{}c@{}}42.95\%\\ (39.59\%,46.22\%)\end{tabular}}  \\
                                           & UMLS\_Recall     &        \begin{tabular}[c]{@{}c@{}}2.82\% \\ (2.39\%,3.34\%)\end{tabular}              &  \begin{tabular}[c]{@{}c@{}}0.08\% \\ (0.01\%,0.15\%)\end{tabular}     & \begin{tabular}[c]{@{}c@{}}2.16\% \\ (1.78\%,2.62\%)\end{tabular}  &  \textbf{\begin{tabular}[c]{@{}c@{}}33.07\%\\ (30.93\%,36.17\%)\end{tabular}}  \\ 
                                           & BERT-Sim & \begin{tabular}[c]{@{}c@{}}39.20\% \\ (38.55\%,40.02\%)\end{tabular}  & \begin{tabular}[c]{@{}c@{}}29.14\% \\ (28.48\%,29.81\%)\end{tabular}  & \begin{tabular}[c]{@{}c@{}}44.72\% \\ (43.97\%,45.65\%)\end{tabular}  & \textbf{\begin{tabular}[c]{@{}c@{}}68.47\% \\ (66.85\%,70.05\%)\end{tabular}}\\ \bottomrule
\end{tabular}
\end{table}

\begin{figure}[t]
    \centering
    \includegraphics[width = \textwidth]{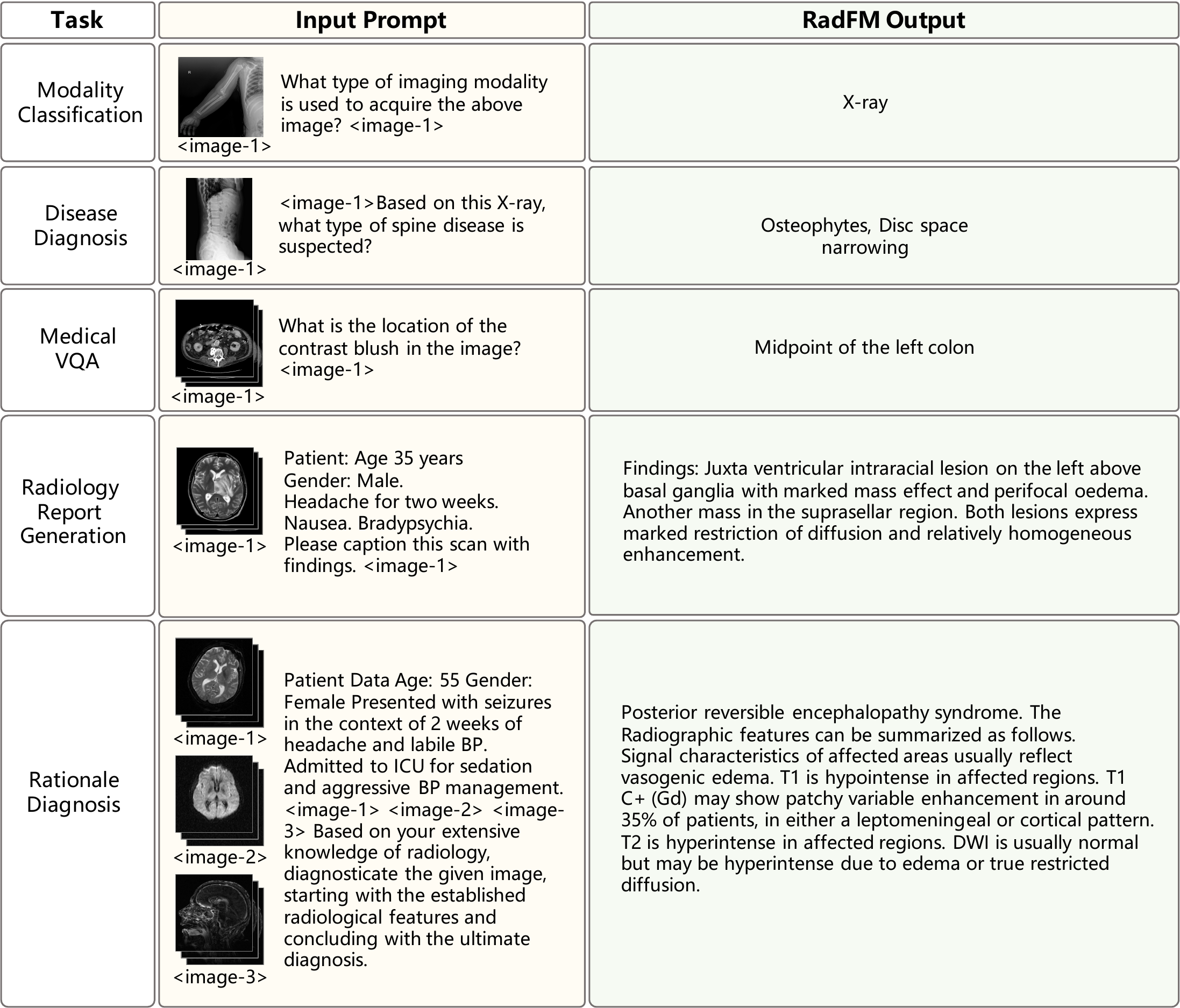}
    \vspace{4pt}
    \caption{Examples of inputs and outputs of five different evaluation tasks obtained from RadFM.}
    \label{fig:task_view}
\end{figure}

In this section,  we start by presenting the evaluation results on RadBench, 
including five different tasks~(Fig.~\ref{fig:task_view}), namely, modality recognition, disease diagnosis, report generation, and visual question-answering on radiologic modalities and anatomical regions. 
\change{Specifically, we compare the zero-shot evaluation results of RadFM with foundation model baselines on both zero-shot (Tab.~\ref{tab_zeroshot_1} and Tab.~\ref{tab_zeroshot_2}) and few-shot (Tab.~\ref{tab_fewshot_1} and Tab.~\ref{tab_fewshot_2}) settings.
Following that, we perform fine-tuning experiments (Tab.~\ref{tab_finetune}) to thoroughly evaluate the performance of our model. Additionally, to evaluate the model's generalization ability, we employed a zero-shot evaluation on the unseen classes in the PadChest dataset (Fig.~\ref{fig:padchest}). It is worth noting that PadChest had not been utilized in the model's training process.}

\begin{table}[!t]
\footnotesize
\caption{\change{Comparison of \textbf{zero-shot} evaluation of the proposed RadFM with foundation model baselines on RadBench across all datasets and metrics. 
The benchmark includes 5 tasks, modality recognition, disease diagnosis, medical visual question answering, report generation, and rationale diagnosis.  We show the results of modality recognition, disease diagnosis, medical visual question answering in this table. ACC, F1, BLEU, ROUGE, $\text{UMLS}\_\text{Precision}$, $\text{UMLS}\_\text{Recall}$ and BERT-Sim are reported, and the metrics refer to the average score on all test samples. Numbers within parentheses indicate $95\%$ CI.
}}
\vspace{5pt}
\centering
\setlength{\tabcolsep}{0.4pt}
\renewcommand{\arraystretch}{1.2}
\begin{tabular}{l|l|l|l|l|l}
\toprule
\rowcolor{lightgray}
Dataset & Metric & OpenFlamingo & MedVInT & Med-Flamingo & RadFM \\
\midrule
\rowcolor{lightgray!50}
\multicolumn{6}{l}{\textbf{Modality Recognition} } \\
RP3D-Modality & ACC & 49.91 (48.07, 50.72) & 84.25 (83.18, 85.39) & 32.87 (31.09, 34.22) & \textbf{92.95 (91.15, 94.6)} \\
\midrule
\rowcolor{lightgray!50}
\multicolumn{6}{l}{\textbf{Disease Diagnosis}} \\
VinDr-Mammo & ACC & 49.62 (47.9, 51.34) & 50.06 (48.52, 51.59) & 49.96 (48.45, 51.47) & \textbf{59.96 (58.41, 61.59)} \\
 & F1 & 28.56 (26.22, 30.9) & 66.56 (65.2, 67.93) & \textbf{66.51 (65.15, 67.87)} & 62.11 (60.09, 63.75) \\
VinDr-SpineXr & ACC & 51.8 (48.77, 54.84) & 49.93 (46.99, 52.86) & 50.69 (47.58, 53.79) & \textbf{68.82 (65.92, 71.47)} \\
 & F1 & 23.79 (19.6, 27.98) & 62.32 (59.38, 65.25) & 65.11 (62.24, 67.98) & \textbf{67.69 (64.5, 70.98)} \\
VinDr-PCXR & ACC & 51.39 (46.46, 56.32) & 50.29 (45.88, 54.69) & 50.43 (45.69, 55.17) & \textbf{56.32 (51.82, 61.21)} \\
 & F1 & 10.33 (4.71, 15.95) & \textbf{66.29 (62.36, 70.23)} & 64.68 (60.27, 69.08) & 37.53 (28.88, 43.67) \\
CXR-Mix & ACC & 50.55 (50.07, 51.03) & 49.2 (48.53, 49.88) & 50.17 (49.66, 50.67) & \textbf{83.62 (83.23, 83.97)} \\
 & F1 & 24.83 (24.11, 25.54) & 67.22 (66.62, 67.82) & 66.11 (65.72, 66.61) &\textbf{82.99 (82.58, 83.49)} \\
RadChest-CT & ACC & 49.66 (47.95, 51.38) & 50.07 (47.68, 52.45) & 49.93 (48.21, 51.64) & \textbf{72.95 (71.06, 74.78)} \\
 & F1 & 13.89 (11.47, 16.31) & 66.57 (64.45, 68.69) & 65.57 (64.03, 67.11) & \textbf{71.86 (69.42, 83.49)} \\
 \midrule
\rowcolor{lightgray!50}
\multicolumn{6}{l}{\textbf{Medical VQA}} \\
PMC-VQA & BLEU & 5.75 (4.33, 6.95) & \textbf{23.73 (21.03, 26.73)} & 7.36 (5.77, 9.03) & 17.99 (14.8, 20.83) \\
 & ROUGE & 10.08 (7.93, 12.15) & \textbf{27.24 (24.04, 30.91)} & 11.67 (9.54, 13.87) & 19.43 (16.56, 23.55) \\
 & UMLS\_Precision & 4.15 (2.43, 6.09) & 19.64 (16.2, 23.59) & 4.86 (3.03, 6.8) & \textbf{20.74 (17.39, 24.71)} \\
 & UMLS\_Recall & 4.66 (2.83, 6.67) & \textbf{18.88 (15.51, 22.68)} & 4.96 (3.28, 6.79) & 14.14 (11.19, 17.37) \\
 & BERT-Sim & 44.19 (42.95, 45.83) & 57.81 (55.49, 59.76) & 42.6 (41.25, 44.08) & \textbf{63.85 (62.04, 65.94)} \\
PMC-CaseReport & BLEU & 21.1 (18.93, 23.74) & 5.68 (4.43, 6.66) & 15.31 (13.64, 17.15) & \textbf{40.48 (37.95, 42.61)} \\
 & ROUGE & 28.06 (25.56, 30.91) & 10.33 (8.67, 11.87) & 20.86 (18.48, 23.18) & \textbf{48.14 (45.6, 50.98)} \\
 & UMLS\_Precision & 27.9 (23.63, 32.29) & 13.38 (10.2, 16.59) & 21.58 (18.17, 24.87) & \textbf{41.59 (37.76, 44.97)} \\
 & UMLS\_Recall & 21.41 (18.17, 24.8) & 5.95 (4.08, 7.77) & 14.81 (12.71, 17.11) & \textbf{40.03 (36.68, 43.53)} \\
 & BERT-Sim & 54.84 (53.28, 56.85) & 42.36 (41.2, 43.97) & 50.18 (48.53, 52.17) &\textbf{72.68 (70.92, 74.94)} \\
VQA-RAD & BLEU & 5.09 (2.72, 10.03) & 35.1 (28.44, 41.55) & 3.21 (1.64, 5.07) & \textbf{52.24 (44.97, 59.43)} \\
 & ROUGE & 9.32 (5.51, 14.98) & 39.2 (31.36, 46.33) & 6.49 (3.16, 9.82) & \textbf{52.74 (45.39, 61.05)} \\
 & UMLS\_Precision & 1.77 (0.16, 3.81) & 16.46 (7.83, 25.93) & 1.6 (0.16, 4.09) & \textbf{62.12 (54.01, 71.12)} \\
 & UMLS\_Recall & 4.91 (0.31, 11.4) & 15.94 (7.72, 25.48) & 2.34 (0.23, 6.5) & \textbf{42.82 (32.31, 51.54)} \\
 & BERT-Sim & 41.13 (38.57, 44.04) & 71.39 (66.94, 75.46) & 40.71 (38.95, 42.79) & \textbf{81.52 (77.41, 85.17)} \\
SLAKE & BLEU & 5.71 (3.49, 8.11) & 24.81 (20.23, 30.52) & 3.83 (2.29, 6.04) & \textbf{78.56 (72.2, 83.28)} \\
 & ROUGE & 13.39 (9.31, 18.61) & 29.08 (24.06, 34.8) & 8.68 (5.15, 13.08) &\textbf{79.42 (75.15, 84.05)} \\
 & UMLS\_Precision & 3.02 (1.16, 5.57) & 23.32 (18.08, 29.42) & 1.57 (0.4, 2.8) & \textbf{81.5 (76.81, 86.87)} \\
 & UMLS\_Recall & 4.49 (1.86, 7.94) & 23.74 (18, 30.08) & 3.18 (0.79, 5.61) & \textbf{74.42 (66.7, 81.19)} \\
 & BERT-Sim & 46.74 (44.86, 48.66) & 67.7 (64.94, 70.69) & 46.04 (44.44, 48) & \textbf{93.30 (90.99, 95.60)} \\
RP3D-VQA & BLEU & 6.42 (6.1, 6.67) & 1.56 (1.31, 1.97) & 15.55 (14.71, 16.44) &\textbf{23.23 (22.16, 24.26)} \\
 & ROUGE & 28.97 (27.79, 30.12) & 3.95 (3.54, 4.51) & 20.56 (19.63, 21.66) & \textbf{30.88 (29.84, 32.16)} \\
 & UMLS\_Precision & 17.4 (16.2, 18.61) & 8.62 (6.8, 10.33) & 21.93 (20.48, 23.66) &\textbf{22.89 (21.02, 24.48)} \\
 & UMLS\_Recall & \textbf{19.78 (18.54, 20.92)} & 1.95 (1.47, 2.46) & 12.98 (11.86, 14.11) & 17.8 (16.43, 19.08) \\
 & BERT-Sim & 46.17 (45.55, 46.66) & 71.39 (66.94, 75.46) & 52.12 (51.37, 52.81) & \textbf{72.13 (70.36, 74.23)} \\
\bottomrule
\end{tabular}
\label{tab_zeroshot_1}
\end{table}

\begin{table}[!tb]
\footnotesize
\caption{
\change{Comparison of \textbf{zero-shot} evaluation of proposed RadFM with foundation model baselines on RadBench across all datasets and metrics. 
The benchmark includes 5 tasks, modality recognition, disease diagnosis, medical visual question answering, report generation, and rationale diagnosis.  We show the results of report generation rationale diagnosis in this table. BLEU, ROUGE, $\text{UMLS}\_\text{Precision}$, $\text{UMLS}\_\text{Recall}$ and BERT-Sim are reported, and the metrics refer to the average score on all test samples. Numbers within parentheses indicate $95\%$ CI.}
}
\vspace{5pt}
\centering
\setlength{\tabcolsep}{1pt}
\renewcommand{\arraystretch}{1.2}
\begin{tabular}{l|l|l|l|l|l}
\toprule
\rowcolor{lightgray}
Dataset & Metric & OpenFlamingo & MedVInT & Med-Flamingo & RadFM \\
\midrule
\rowcolor{lightgray!50}
\multicolumn{6}{l}{\textbf{Report Generation }} \\
MIMIC-CXR & BLEU & 5.11 (4.14, 6.16) & 0.04 (0.01, 0.08) & 10.47 (9.45, 11.66) & \textbf{19.43 (16.12, 23.25)} \\
 & ROUGE & 6.32 (5.02, 7.87) & 2.69 (2.26, 3.15) & 10.38 (9.7, 11.27) & \textbf{26.18 (23.07, 29.86)} \\
 & UMLS\_Precision & 3.68 (2.18, 6.11) & 26.67 (11.19, 42.12) & 6.57 (5.09, 8.09) & \textbf{45.51 (40.47, 52.77)} \\
 & UMLS\_Recall & 1.13 (0.67, 1.67) & 0.52 (0.2, 0.88) & 2.34 (1.75, 2.86) & \textbf{23.39 (20.18, 27.53)} \\
 & BERT-Sim & 40.87 (39.39, 42.67) & 34.48 (32.69, 36.02) & 48.96 (48.03, 49.91) &  \textbf{66.77 (64.87, 68.58)} \\
RP3D-Caption & BLEU & 3.25 (2.24, 4.23) & 1.52 (1.29, 1.82) & 9.91 (9.4, 10.37) & \textbf{10.21 (9.48, 11.03)} \\
 & ROUGE & 7.17 (5.26, 9.24) & 4.84 (4.44, 5.23) & \textbf{15.62 (14.96, 16.17)} & 15.51 (14.79, 16.26) \\
 & UMLS\_Precision & 1.13 (0.18, 2.78) & 8.38 (6.36, 10.4) & 2.57 (2.14, 3.06) & \textbf{18.97 (18.12, 19.98)} \\
 & UMLS\_Recall & 1.35 (0.19, 3.3) & 1.12 (0.85, 1.41) & 2.03 (1.63, 2.4) & \textbf{9.32 (8.81, 9.89)} \\
 & BERT-Sim & 37.18 (35.76, 38.45) & 40.67 (39.86, 41.56) & 47.98 (47.6, 48.37) & \textbf{56.78 (56.40, 57.22)} \\
MedPix-single-caption & BLEU & 4.82 (3.73, 6.11) & 0.89 (0.18, 1.86) & 10.37 (8.15, 12.45) & \textbf{12.72 (7.95, 21.32)} \\
 & ROUGE & 9.59 (8.06, 11.21) & 3.29 (2.11, 4.46) & 16.77 (14.24, 19.83) & \textbf{17.84 (12.32, 25.57)} \\
 & UMLS\_Precision & 2.58 (1.36, 4.5) & 3.04 (0, 8.67) & 2.23 (1.29, 3.51) & \textbf{12.42 (7.69, 18.46)} \\
 & UMLS\_Recall & 1.17 (0.43, 2.1) & 0.39 (0, 1.29) & 3.3 (1.36, 5.52) & \textbf{9.15 (4.09, 13.93)} \\
 & BERT-Sim & 40.22 (39.34, 41.19) & 33.04 (30.38, 35.67) & 47.52 (45.95, 49.15) &  \textbf{57.94 (56.13, 58.99)} \\
MedPix-multi-caption  & BLEU & 4.34 (4.03, 4.62) & 0.88 (0.38, 1.77) & 9.19 (7.51, 10.77) & \textbf{21.59 (16.51, 28.79)} \\
 & ROUGE & 6.19 (5.91, 6.58) & 2.82 (2.09, 3.86) & 14.57 (12.33, 16.49) & \textbf{26.25 (20.87, 32.67)} \\
 & UMLS\_Precision & 1.38 (0.93, 1.92) & 2.87 (0, 8.18) & 3.21 (1.51, 5.94) & \textbf{19.11 (14.37, 25.78)} \\
 & UMLS\_Recall & 0.37 (0.24, 0.58) & 0.19 (0, 0.65) & 1.88 (0.91, 3.34) & \textbf{16.59 (11.75, 23.63)} \\
 & BERT-Sim & 39.97 (39.78, 40.23) & 33.48 (32.02, 35.3) & 46.56 (44.82, 47.97) &  \textbf{61.27 (58.38, 64.60)} \\
 \midrule
\rowcolor{lightgray!50}
\multicolumn{6}{l}{\textbf{Rationale Diagnosis}} \\
RP3D-Ration & BLEU & 3.63 (4.88, 4.12) & 0.09 (0.01, 0.17) & 7.65 (7, 8.37) & \textbf{34.6 (31.69, 37.74)} \\
 & ROUGE & 4.1 (4.98, 4.56) & 0.67 (0.52, 0.83) & 7.38 (6.86, 8.01) & \textbf{41.89 (39.2, 44.77)} \\
 & UMLS\_Precision & 8.49 (14.9, 11.58) & 9.07 (1.91, 16.23) & 5.97 (4.82, 7.08) & \textbf{42.95 (39.59, 46.22)} \\
 & UMLS\_Recall & 0.66 (1.29, 0.96) & 0.08 (0.01, 0.15) & 2.17 (1.78, 2.66) & \textbf{33.07 (30.93, 36.17)} \\
 & BERT-Sim & 38.55 (40.02, 39.2) & 29.14 (28.48, 29.81) & 44.72 (43.97, 45.65) &  \textbf{68.47 (66.85, 70.05)} \\
\bottomrule
\end{tabular}
\label{tab_zeroshot_2}
\end{table}

\begin{table}[!tb]
\footnotesize
\caption{\change{Comparison of \textbf{zero-shot} evaluation of RadFM with \textbf{few-shot} evaluation of OpenFlamingo and Med-Flamingo.
The benchmark includes 5 tasks, modality recognition, disease diagnosis, medical visual question answering, report generation, and rationale diagnosis. We show the results of modality recognition, disease diagnosis, medical visual question answering in this table.ACC, F1, BLEU, ROUGE, $\text{UMLS}\_\text{Precision}$, $\text{UMLS}\_\text{Recall}$ and BERT-Sim are reported, and the metrics refer to the average score on all test samples. Numbers within parentheses indicate $95\%$ CI.}
}
\vspace{5pt}
\centering
\setlength{\tabcolsep}{3pt}
\renewcommand{\arraystretch}{1.2}
\begin{tabular}{l|l|l|l|l}
\toprule
\rowcolor{lightgray}
Dataset & Metric & OpenFlamingo (few-shot) & Med-Flamingo (few-shot) & RadFM\\
 \midrule
 \rowcolor{lightgray!50}
\multicolumn{5}{l}{\textbf{Modality Recognition}} \\
RP3D-Modality & ACC & 57.06 (55.65, 58.65) & 47.68 (46.48, 49.07) & \textbf{92.95 (91.15, 94.6)} \\
 \midrule
\rowcolor{lightgray!50}
\multicolumn{5}{l}{\textbf{Disease Diagnosis}} \\
VinDr-Mammo & ACC & 49.92 (48.2, 51.65) & 49.88 (48.33, 51.42) & \textbf{59.96 (58.41, 61.59)} \\
& F1 & 57.01 (55.3, 58.72) & \textbf{64.92 (63.52, 66.32)} & 62.11 (60.09, 63.75) \\
VinDr-SpineXr & ACC & 50.33 (47.13, 53.53) & 49.61 (46.05, 53.16) & \textbf{68.82 (65.92, 71.47)}\\
& F1 & 31.79 (26.99, 36.58) & 63.23 (59.74, 66.74) & \textbf{67.69 (64.5, 70.98)} \\
VinDr-PCXR & ACC & 49.85 (45.4, 54.31) & 49.37 (44.44, 54.31) & \textbf{56.32 (51.82, 61.21)} \\
& F1 & 41.43 (33.77, 49.1) & \textbf{66.94 (62.57, 71.32)} & 37.53 (28.88, 43.67) \\
CXR-Mix & ACC & 50.63 (50.07, 51.17) & 50 (49.5, 50.51) & \textbf{83.62 (83.23, 83.97)} \\
& F1 & 65.84 (65.34, 66.35) & 66.66 (66.2, 67.1) & \textbf{82.99 (82.58, 83.49)}\\
RadChest-CT & ACC & 50.93 (49.13, 52.72) & 50.39 (48.34, 52.43) & \textbf{72.95 (71.06, 74.78)} \\
& F1 & 43.49 (41.18, 45.99) & 63.31 (61.39, 65.23) & \textbf{71.86 (69.42, 83.49)} \\
 \midrule
\rowcolor{lightgray!50}
\multicolumn{5}{l}{\textbf{Medical VQA}} \\
PMC-VQA & BLEU & 11.1 (8.93, 13.41) & 11.03 (9.24, 13.49) & \textbf{17.99 (14.8, 20.83)} \\
& ROUGE & 13.03 (10.63, 15.46) & 13.06 (10.93, 15.66) & \textbf{19.43 (16.56, 23.55)} \\
& UMLS\_Precision & 7.6 (5.41, 10.83) & 6.35 (4.05, 8.97) & \textbf{20.74 (17.39, 24.71)} \\
& UMLS\_Recall & 7.56 (5.4, 10.51) & 6.1 (4.04, 8.71) & \textbf{14.14 (11.19, 17.37)} \\
& BERT-Sim & 52.08 (50.43, 54.07) & 51.37 (49.57, 53.01) & \textbf{63.85 (62.04, 65.94)} \\
PMC-CaseReport & BLEU & 29.75 (27.35, 31.85) & 26.14 (23.78, 28.36) & \textbf{40.48 (37.95, 42.61)} \\
& ROUGE & 37.65 (35.06, 39.86) & 33.03 (30.52, 35.16) & \textbf{48.14 (45.6, 50.98)} \\
& UMLS\_Precision & 33.59 (30.12, 37.16) & 34.34 (31.13, 37.21) & \textbf{41.59 (37.76, 44.97)} \\
& UMLS\_Recall & 30.74 (26.96, 34.02) & 26.05 (22.67, 28.74) & \textbf{40.03 (36.68, 43.53)} \\
& BERT-Sim & 61.3 (59.57, 62.94) & 59.17 (57.5, 60.49) &  
\textbf{72.68 (70.92, 74.94)} \\
VQA-RAD & BLEU & 33.98 (26.75, 42.52) & 35.97 (29.14, 45.45) & \textbf{52.24 (44.97, 59.43)} \\
& ROUGE & 35.26 (28.21, 43.91) & 38.64 (31.42, 48.23) & \textbf{52.74 (45.39, 61.05)} \\
& UMLS\_Precision & 14.72 (6.86, 24.22) & 18.7 (8.99, 29.61) &\textbf{62.12 (54.01, 71.12)} \\
& UMLS\_Recall & 14.52 (7.63, 23.33) & 17.46 (8.76, 27.85) & \textbf{42.82 (32.31, 51.54) }\\
& BERT-Sim & 71.49 (67.63, 74.96) & 73.4 (69.62, 77.32) &  
\textbf{81.52 (77.41, 85.17)} \\
SLAKE & BLEU & 27.16 (22.01, 32.56) & 23.62 (18.06, 28.26) & \textbf{78.56 (72.2, 83.28)} \\
& ROUGE & 29.36 (24.23, 34.73) & 24.86 (19.47, 29.94) & \textbf{79.42 (75.15, 84.05)} \\
& UMLS\_Precision & 23.58 (17.52, 30.73) & 18.28 (13.23, 23.38) & \textbf{81.5 (76.81, 86.87)} \\
& UMLS\_Recall & 22.71 (17.48, 29.53) & 19.21 (13.38, 24.37) &\textbf{74.42 (66.7, 81.19)} \\
& BERT-Sim & 69.42 (66.09, 72.04) & 66.93 (63.98, 70.32) & \textbf{93.30 (90.99, 95.60)} \\
RP3D-VQA & BLEU & 19.93 (18.73, 21.07) & 18.68 (17.77, 19.78) & \textbf{23.23 (22.16, 24.26)} \\
& ROUGE & 26.27 (24.83, 27.55) & 24.86 (23.86, 26.15) & \textbf{30.88 (29.84, 32.16)} \\
& UMLS\_Precision & 23.28 (21.42, 25.19) & 19.42 (17.75, 21.03) & \textbf{22.89 (21.02, 24.48)} \\
& UMLS\_Recall & 17.19 (15.73, 18.62) & 14.19 (12.84, 15.55) & \textbf{17.8 (16.43, 19.08)} \\
& BERT-Sim & 58.24 (57.59, 58.97) & 57.34 (56.64, 58.09) & \textbf{72.13 (70.36, 74.23)} \\
\bottomrule
\end{tabular}
\label{tab_fewshot_1}
\end{table}

\begin{table}[!tb]
\footnotesize
\caption{\change{Comparison of \textbf{zero-shot} evaluation of RadFM with \textbf{few-shot} evaluation of OpenFlamingo and Med-Flamingo.
The benchmark includes 5 tasks, modality recognition, disease diagnosis, medical visual question answering, report generation, and rationale diagnosis. We show the results of report generation rationale diagnosis in this table. BLEU, ROUGE, $\text{UMLS}\_\text{Precision}$, $\text{UMLS}\_\text{Recall}$ and BERT-Sim are reported, and the metrics refer to the average score on all test samples. Numbers within parentheses indicate $95\%$ CI.}
}
\vspace{5pt}
\centering
\setlength{\tabcolsep}{3pt}
\renewcommand{\arraystretch}{1.2}
\begin{tabular}{l|l|l|l|l}
\toprule
\rowcolor{lightgray}
Dataset & Metric & OpenFlamingo (few-shot) & Med-Flamingo (few-shot) & RadFM\\
 \midrule
\rowcolor{lightgray!50}
\multicolumn{5}{l}{\textbf{Report Generation }} \\
MIMIC-CXR & BLEU & \textbf{23.79 (22.62, 24.86) }& 22.65 (20.93, 24.06) & 19.43 (16.12, 23.25) \\
& ROUGE & \textbf{35.83 (33.7, 37.96)} & 27.29 (25.63, 29.04) & 26.18 (23.07, 29.86) \\
& UMLS\_Precision & 16.75 (15.74, 17.88) & 22.36 (20.91, 23.81) & \textbf{45.51 (40.47, 52.77)} \\
& UMLS\_Recall & \textbf{24.93 (22.86, 27.38)} & 19.64 (17.89, 21.43) & 23.39 (20.18, 27.53) \\
& BERT-Sim & 65.91 (65.2, 66.7) & 66.03 (65.37, 66.83) &  \textbf{66.77 (64.87, 68.58)} \\
RP3D-Caption & BLEU & 1.94 (1.3, 2.71) & 4.97 (4.53, 5.4) & \textbf{10.21 (9.48, 11.03)} \\
& ROUGE & 3.59 (2.34, 5.07) & 6.96 (6.32, 7.44) & \textbf{15.51 (14.79, 16.26)} \\
& UMLS\_Precision & 0.77 (0, 2.23) & 2 (1.58, 2.51) & \textbf{18.97 (18.12, 19.98)} \\
& UMLS\_Recall & 0.71 (0, 2.26) & 1.15 (0.87, 1.41) & \textbf{9.32 (8.81, 9.89)} \\
& BERT-Sim & 36.01 (34.42, 37.1) & 44.98 (44.26, 45.7) & \textbf{56.78 (56.40, 57.22)} \\
MedPix-single-caption  & BLEU & 5.42 (4.42, 6.62) & 8.97 (7.13, 10.64) & \textbf{12.72 (7.95, 21.32)} \\
& ROUGE & 12.63 (10.92, 14.32) & 14.48 (11.67, 17.67) & \textbf{17.84 (12.32, 25.57)} \\
& UMLS\_Precision & 2.61 (1.42, 3.89) & 1.49 (0.44, 2.58) & \textbf{12.42 (7.69, 18.46)} \\
& UMLS\_Recall & 1.65 (0.85, 2.54) & 2.28 (0.54, 4.64) & \textbf{9.15 (4.09, 13.93)} \\
& BERT-Sim & 43.41 (42.37, 44.35) & 50.4 (48.63, 52) & \textbf{57.94 (56.13, 58.99)} \\
MedPix-multi-caption  & BLEU & 7.37 (6.9, 7.88) & 8.83 (7.28, 10.34) & \textbf{21.59 (16.51, 28.79)} \\
& ROUGE & 10.27 (9.81, 10.78) & 13.03 (11.29, 15.14) & \textbf{26.25 (20.87, 32.67)} \\
& UMLS\_Precision & 1.5 (1.15, 1.89) & 4.67 (2.63, 6.5) &\textbf{ 19.11 (14.37, 25.78)} \\
& UMLS\_Recall & 1.56 (1.25, 1.84) & 1.94 (1.19, 2.96) & \textbf{16.59 (11.75, 23.63)} \\
& BERT-Sim & 49.03 (48.44, 49.71) & 47.83 (46.63, 49.06) & \textbf{61.27 (58.38, 64.60)} \\
\midrule
\rowcolor{lightgray!50}
\multicolumn{5}{l}{\textbf{Rationale Diagnosis}} \\
RP3D-Ration & BLEU & 17.52 (18.86, 18.1) & 17.15 (17.81, 17.15) & \textbf{34.6 (31.69, 37.74)} \\
& ROUGE & 28.4 (30.88, 29.63) & 29.02 (30.28, 29.02) & \textbf{41.89 (39.2, 44.77)} \\
& UMLS\_Precision & 18.14 (20.51, 19.26) & 21.04 (22.19, 21.04) & \textbf{42.95 (39.59, 46.22)} \\
& UMLS\_Recall & 15.55 (17.86, 16.68) & 16.89 (18.24, 16.89) & \textbf{33.07 (30.93, 36.17)} \\
& BERT-Sim & 53.61 (54.57, 54.11) & 54.38 (54.89, 54.38) & \textbf{68.47 (66.85, 70.05)} \\
\bottomrule
\end{tabular}
\label{tab_fewshot_2}
\end{table}

\begin{figure}[!htb]
    \centering
    \includegraphics[width = \textwidth]{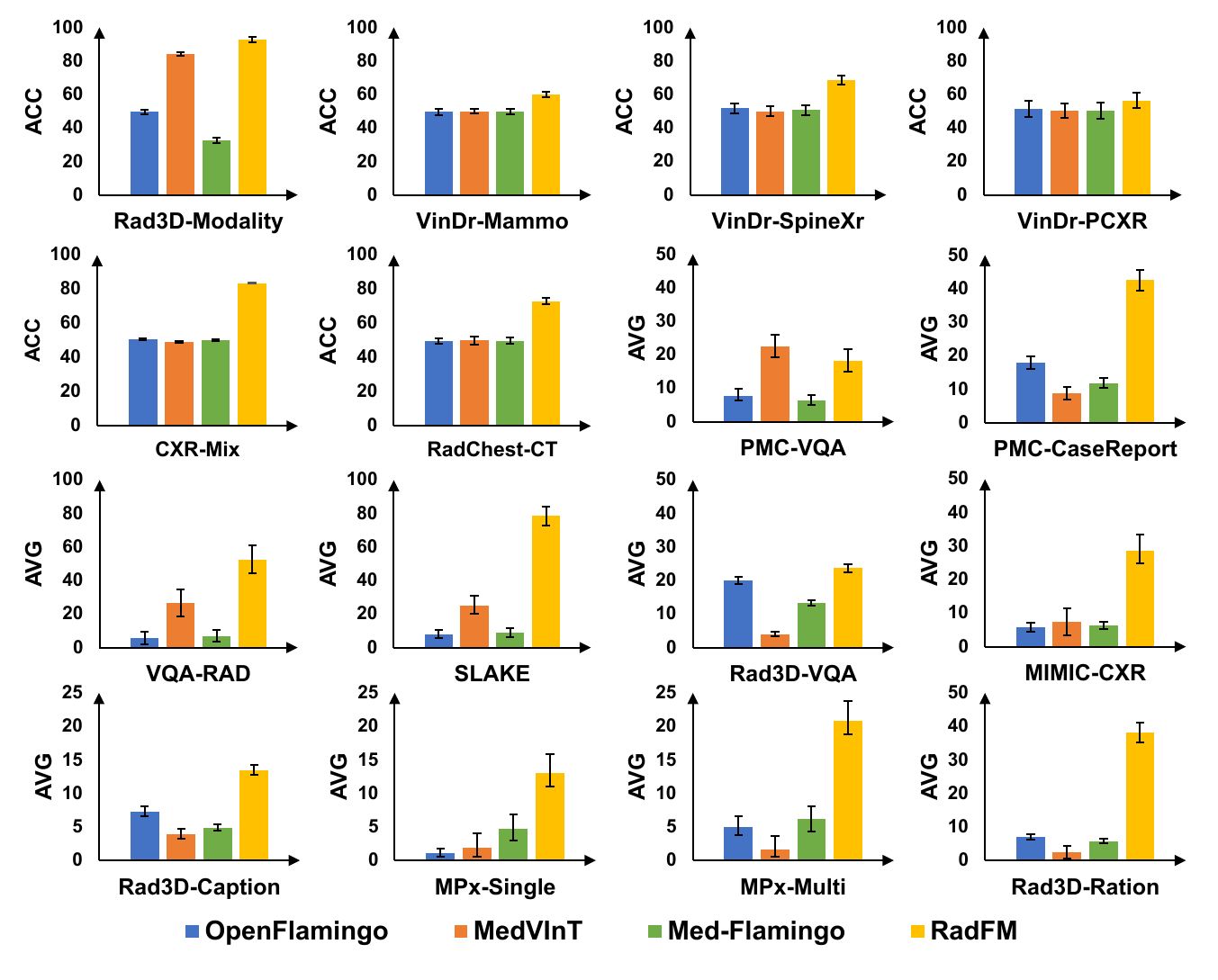}
    \caption{Comparison of RadFM with various foundation models on different subsets. 
    Except from task level, we also report detailed results on each combined dataset.
    For the dataset involving modality recognition and diagnosis, like RP3D-Modality, VinDr-Mammo, VinDr-SpineX, VinDr-PCXR, CXR-Mix, RadChest-CT, ACC scores are plotted in the figure. For the left datasets, AVG scores, denoting the average of the \change{four word-overlap based metrics evaluation metrics}, \emph{i.e.}, BLEU, ROUGE, UMLS\_Precision and UMLS\_Recall, are plotted and the error bars are based on $95\%$ CI.}
    \label{fig:result_dataset}
\end{figure}

\subsection{Results on RadBench}

\subsubsection{Modality Recognition}
Recognizing image modality is one basic skill expected of any advanced foundation models, however, as depicted in Tab.~\ref{tab_zeroshot_1}, 
the foundation models like OpenFlamingo and Med-Flamingo do not perform well on this task, with a disappointing accuracy score (ACC) of merely $49.47\%$, highlighting their models cannot distinguish various medical imaging modalities. 
\change{It is worth noting that the few-shot results for both models will be better than zero-shot as shown in Tab.~\ref{tab_fewshot_1}. }In contrast, our proposed RadFM model outperforms the other competitors by a substantial margin. Note that, such task has been formulated as an open-ended question, where the model's output needs to explicitly state the recognized modality as `CT', `MRI', {\em etc}.. 

\subsubsection{Disease Diagnosis}
In clinical practice, the primary and indispensable function of any advanced medical foundation model is to provide assistance in disease diagnosis. 
\change{
As depicted in Tab.~\ref{tab_zeroshot_1} and Tab.~\ref{tab_fewshot_1}, the performance of existing models has been rather poor on such tasks, with accuracy score (ACC) nearly $50\%$. 
}
Considering that we prompt the problem with a judgment format, \emph{i.e.}, ``Does the patient have \{disease\}?'', this score is nearly random. 
In contrast, our model, RadFM, proves its superiority over existing methods with an ACC score of $80.62\%$, making it more suited for real-world clinical applications. Essentially, this resembles a multi-label classification task, with over $5000$ in distinct categories, 
further compounding the challenges of models.

\subsubsection{Medical Visual Question Answering~(VQA)}
In contrast to the aforementioned tasks, 
Medical VQA represents a more comprehensive and versatile challenge in the field of medical image analysis. 
In a clinical setting, patients and radiologists may pose a wide variety of questions related to medical images, ranging from simple inquiries about image modality to more complex reasoning queries.
Here, we combine 5 different medical VQA datasets for evaluation,
namely, PMC-VQA, PMC-CaseReport, VQA-RAD, SLAKE and RP3D-VQA.

As shown in Tab.~\ref{tab_zeroshot_1} and Fig.~\ref{fig:result_dataset}, comparing to the second best model, MedVInT, which was specifically trained on visual question answering, despite achieving better results on its in-domain PMC-VQA test set, its generalization is quite poor, even though the task is still medical visual question answering. 
For example, MedVInT struggles on contextual VQA, like PMC-CaseReport, which requires a deeper understanding of the context related to the patient and image, and on real 3D medical scans, like RP3D-VQA, which requires a model capturing the information from an extra image dimension. 
In contrast, our RadFM model shows a substantial improvement in UMLS\_Precision from 14.55\% to 31.77\% and UMLS\_Recall from 14.56\% to 24.93\% across the whole test set, demonstrating its proficiency in comprehensively understanding the given textual information and flexible adaptation to various complex clinical scenarios.

\subsubsection{Report Generation}
Report generation is a crucial and prominent use case for generative medical foundational models. 
Unlike Medical VQA, this application generally requires the model to emphasize clinically significant observations based on the image. 
As shown in Tab.~\ref{tab_zeroshot_2}, RadFM shows significant improvement over existing models, across various metrics, particularly in relation to medical-specific terminology. For instance, RadFM improves UMLS\_Precision from 11.16\% to 22.49\%, and UMLS\_Recall from 2.84\% to 12.07\%.

\subsubsection{Rationale Diagnosis}
In addition to basic diagnosis, the ability to scrutinize diagnostic prediction outcomes is crucial, particularly in light of the stringent demands for precision and interpretability within medical contexts. 
Much like report generation, this task also requires proficiency in generating extended paragraphs and comprehensive understanding on medical knowledge. 

As indicated in Tab.~\ref{tab_zeroshot_1}, 
RadFM is the only model that can effectively respond on this task. 
RadFM outperforms traditional BLEU and ROUGE scores by 8.56\% and 15.46\% respectively. Moreover, it exhibits significant improvements in UMLS\_Precision and UMLS\_Recall scores, showcasing advancements of 23.16\% and 10.96\% respectively. 

\subsubsection{Human Rating}

\begin{figure}[!htb]
    \centering
    \includegraphics[width = \textwidth]{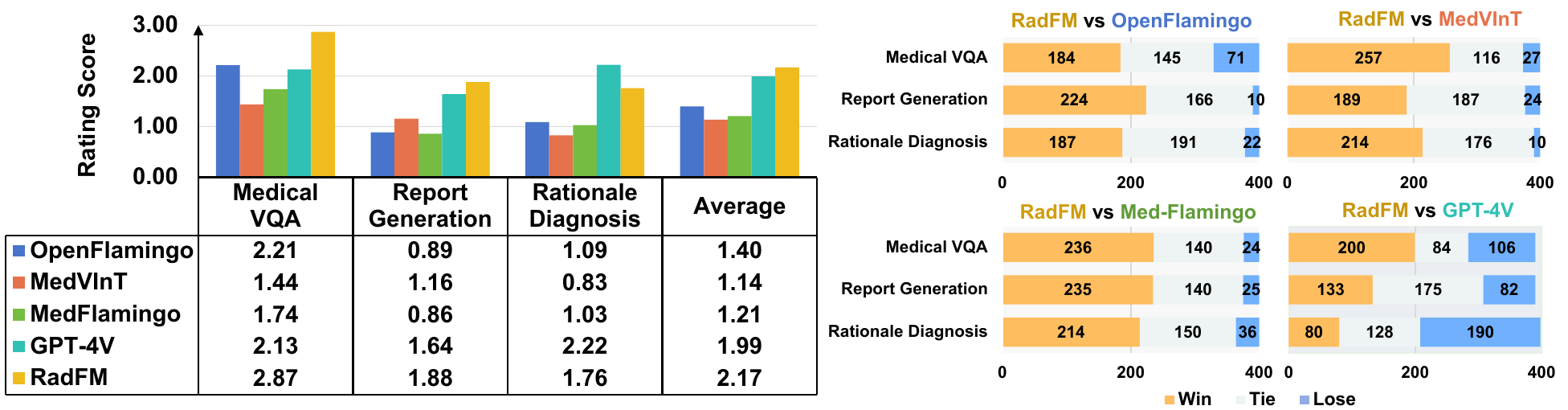}
    \vspace{1pt}
    \caption{Comparison of RadFM with other methods on human rating scores. On the left, we show the absolute human rating scores of different methods on the three generative tasks. \emph{i.e.}, VQA, Report Generation and Rationale Diagnosis. On the right, we show the relative comparison. Each sub-figure in right shows the number of RadFM win/tie/lose cases when comparing against a certain model. \change{Note that, considering GPT-4V may refuse answering medical questions for safety, we dismiss such cases when calculating the scores or comparison relating to GPT-4V. In detail, for 1200 testing cases, 22 cases dismissed for GPT-4V due to safety.}}
    \label{fig:human score}
\end{figure}

In Fig.~\ref{fig:human score}, we show the human rating results of all the models. As shown on the left of the figure, RadFM can achieve higher scores on all three generative-based tasks compared with former methods. On the right, we further show the relative comparison between RadFM and a certain model. In all cases, outputted results from RadFM are preferred by human clinicians. It is worth highlighting that \change{we also show the comparison between RadFM and GPT-4V(ison) which has been widely considered as the strongest foundation model. 
Note that, GPT-4V can only input 4 2D pictures per query, 
we thus ask the radiologists to pick out the most informative 4 slices based on the references answer from 3D volumns. With human prior, the questions will be easier compared with directly inputting original 3D volumns which is used as the evaluation style for our model. 
Despite this, RadFM still surpasses GPT-4V}.

\subsection{Transferring Ability to Public Benchmarks}
\change{In Tab.~\ref{tab_finetune}, 
we treat RadFM as a pre-trained model and finetune on various datasets.
For diagnosis, we use the image-encoder weights as initialization for both 2D and 3D imaging modalities, for VQA and report generation, the whole model are further finetuned on the specific dataset, as shown in the table, our model can improve both diagnosis results and text generation quality, based on the automatic metrics. In general, the representation learned in RadFM benefits various clinical tasks across diverse medical imaging modalities.
}

\begin{table}[!htb]
\footnotesize
\caption{\change{Comparison of RadFM with SOTA models on disease diagnosis, medical visual question answering, report generation. All models were fine-tuned and evaluated on the same train/test set. AUC, F1, BLEU, and ROUGE are reported, and the metrics refer to the average score on all test samples. Numbers within parentheses indicate $95\%$ CI.}
}
\vspace{5pt}
\centering
\setlength{\tabcolsep}{8pt}
\renewcommand{\arraystretch}{1.2}
\begin{tabular}{ll|l|ll}
\toprule
\rowcolor{lightgray}
Dataset & Modality & Metric & SOTA & RadFM \\
 \midrule
\rowcolor{lightgray!50}
\multicolumn{5}{l}{\textbf{Disease Diagnosis}} \\
VinDr-Mammo & 2D Mammography & Macro-AUC & 64.5~\cite{wantlin2023benchmd} & \textbf{64.76 (64.23, 65.88)} \\
 & & Macro-F1 & N/A & \textbf{39.42 (39.37, 39.59)} \\
CXR14 & 2D X-ray & Macro-AUC & 80.1~\cite{Wu2023MedKLIPMK} & \textbf{81.13 (81.07, 81.18)} \\
 & & Macro-F1 & N/A & \textbf{30.20 (30.17, 30.22)} \\
LDCT & 3D CT &  Macro-AUC & 82.1~\cite{wantlin2023benchmd} & \textbf{83.23 (81.97, 85.85)} \\
 & & Macro-F1 & N/A & \textbf{58.34 (57.38, 61.23)} \\
BraTs2019 & 3D MRI &  AUC & 88.06~\cite{chatterjee2022classification} & \textbf{90.61 (85.66, 92.13)} \\
 & & F1 & 90.36~\cite{chatterjee2022classification} & \textbf{92.21 (92.01, 93.21)} \\ 
  \midrule
 \rowcolor{lightgray!50}
\multicolumn{5}{l}{\textbf{Medical VQA}} \\ 
VQA-RAD & 2D Radiology & Bleu & 71.03~\cite{bazi2023vision} & \textbf{73.44 (66.04, 82.18)} \\
 & & Rogue & N/A & \textbf{73.81 (67.80, 80.04)} \\
 & & F1 & N/A & \textbf{78.09 (73.54, 81.90)} \\
SLAKE & 2D Radiology &  Bleu & 78.6~\cite{van2023open} & \textbf{83.16 (79.68, 87.10)} \\
 & & Rogue & N/A & \textbf{83.65 (80.39, 87.10)} \\
 & & F1 & 78.1~\cite{van2023open} & \textbf{84.37 (81.60, 86.78)} \\
PMC-VQA & 2D Radiology & Bleu & 23.69 (20.70, 26.93)~\cite{Zhang2023PMCVQAVI} & \textbf{24.13 (21.01, 27.91)} \\
 & & Rogue & \textbf{27.20 (24.09, 31.13)}~\cite{Zhang2023PMCVQAVI} & 25.64 (22.73, 29.29) \\
 & & F1 & 43.93 (41.16, 46.43)~\cite{Zhang2023PMCVQAVI} & \textbf{48.50 (46.19, 51.00)} \\
  \midrule
 \rowcolor{lightgray!50}
 \multicolumn{5}{l}{\textbf{Report Generation}} \\ 
IU-Xray & 2D X-ray & Bleu-1 & \textbf{38.7}~\cite{ALFARGHALY2021100557} & 37.88 (35.96, 39.32) \\
        &  & Bleu-2 & 24.5~\cite{ALFARGHALY2021100557} & \textbf{24.62 (22.73, 26.94)}\\
        &  & Bleu-3 & 16.6~\cite{ALFARGHALY2021100557} & \textbf{17.72 (15.77, 19.69)} \\
        &  & Bleu-4 & \textbf{11.1}~\cite{ALFARGHALY2021100557} & 10.28 (8.89, 11.64) \\
        &  & Rogue-L & 28.9~\cite{ALFARGHALY2021100557} & \textbf{29.51 (28.09, 30.61)} \\
\bottomrule
\end{tabular}
\label{tab_finetune}
\end{table}

\subsection{Generalization to Unseen Classes in PadChest}
\begin{figure}[!htb]
    \centering
    \includegraphics[width = \textwidth]{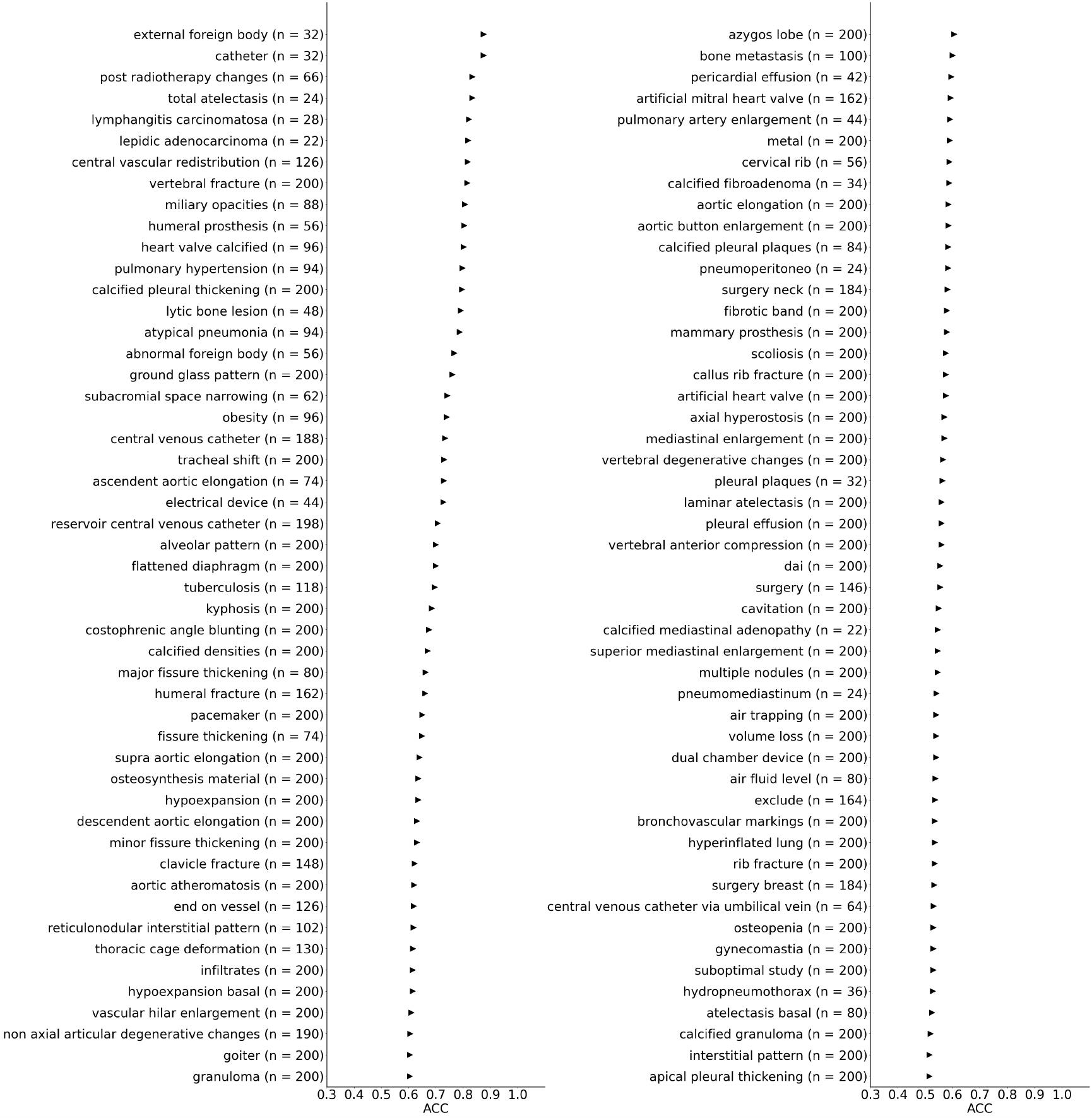}
    \vspace{1pt}
    \caption{\change{Zero-shot evaluation of RadFM on the unseen classes in the PadChest dataset. We evaluate the model on the human-annotated subset of the PadChest dataset, and ACC scores are shown for the radiographic findings or diagnosis. The top 100 classes in the test dataset are shown in the figure. 
}}
    \label{fig:padchest}
\end{figure}

\change{In Fig.~\ref{fig:padchest}, we show the results of zero-shot evaluation of RadFM on the unseen classes in the PadChest dataset. We modify the task as an induction task, for each disease, we randomly select a prompt sentence like ``Is \{disease\} shown in this image'' as input to ask the network to answer whether the case has this disease.
Note that we balance the ratio of `yes' or `no' in the test set and all the disease classes never appeared on the training set. We only plot results from RadFM, as the other foundation model baselines all struggle on this task, getting a random result~(50\% accuracy) for all the classes.}

\subsection{Qualitative Results} 
\change{In this section, we show the qualitative results for different free-form text generation tasks.}

\begin{figure}[!tb]
    \centering
    \includegraphics[width = \textwidth]{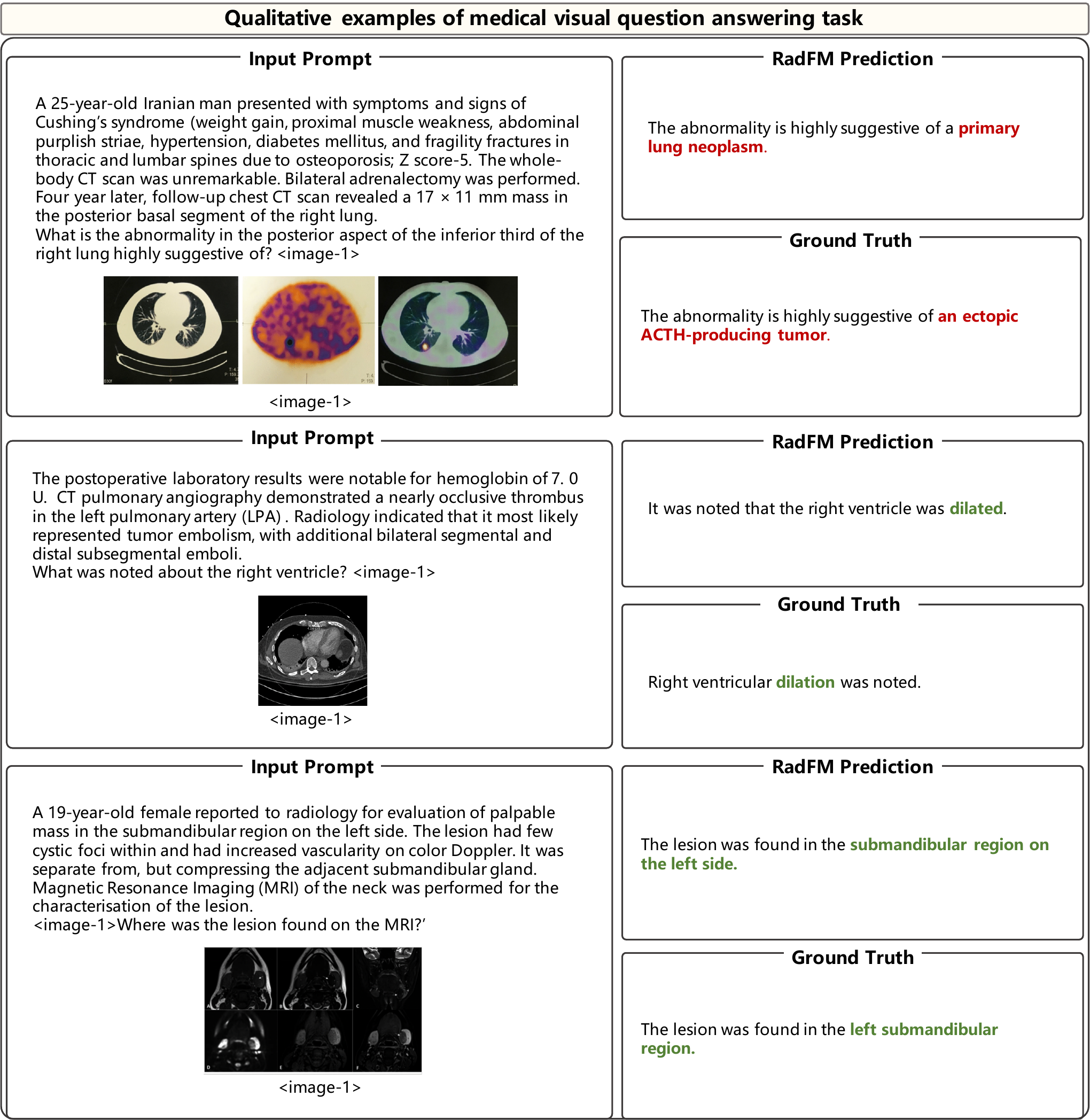}
    \vspace{4pt}
    \caption{Qualitative examples of medical visual question answering (VQA). We present several examples with answers generated by RadFM along with the target ground truth. The green color highlights accurate keywords, while the red color indicates prediction errors. }
    \vspace{4pt}
    \label{fig:result_vqa}
\end{figure}

\change{For medical VQA, Qualitatively, as shown in Fig.~\ref{fig:result_vqa}, 
RadFM demonstrates its ability to comprehend the questions and provide answers in a consistent format, accurately addressing the questions.
However, in some challenging cases, such as the first example where the question pertains to the type of abnormality, the model faces difficulty predicting ``ectopic ACTH-producing tumor'' and mistakenly identifies it as ``primary lung neoplasm'', which requires fine-grained discrimination within tumor types.}


\begin{figure}[!tb]
    \centering
    \includegraphics[width = \textwidth]{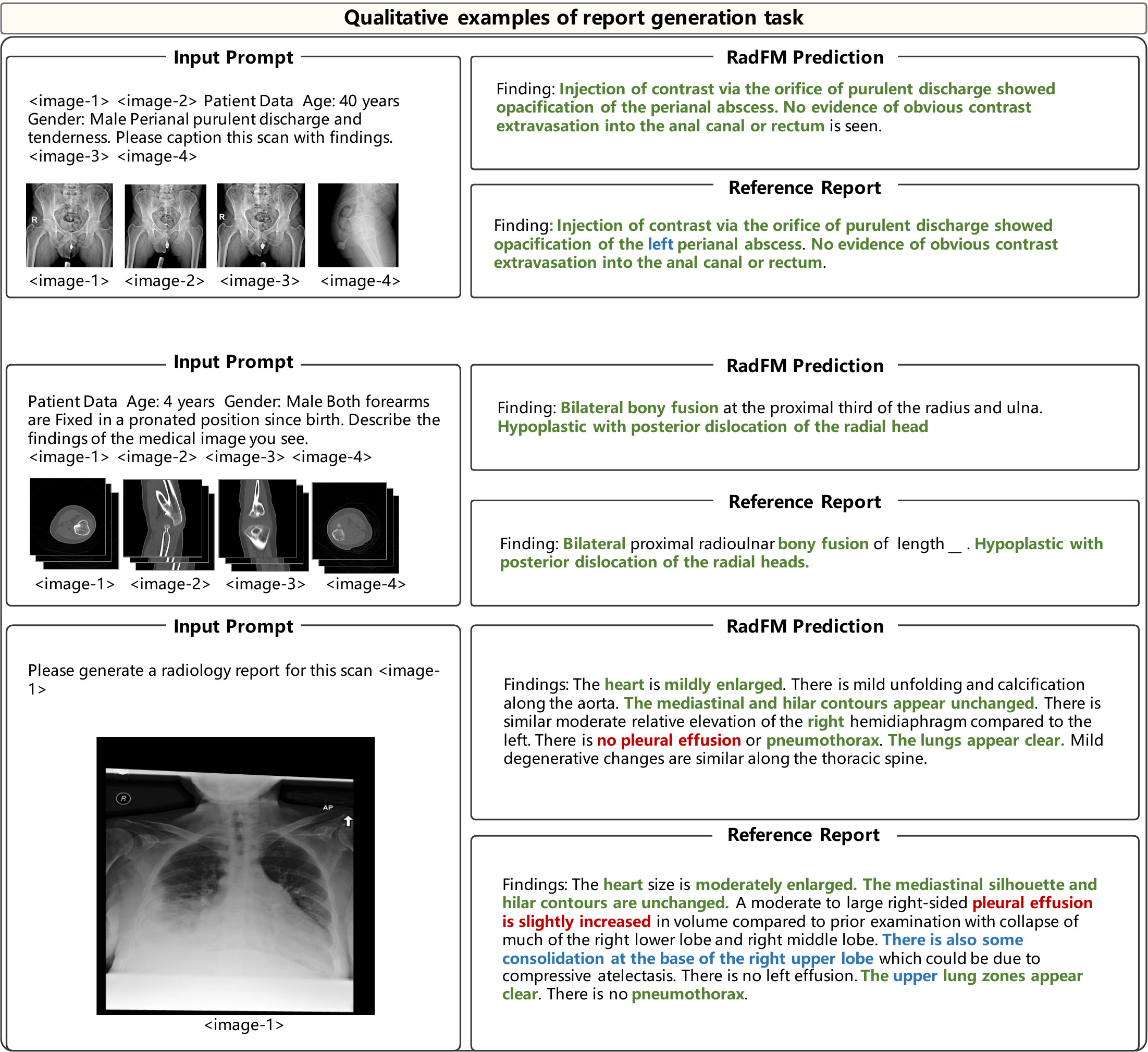}
    \vspace{4pt}
    \caption{Qualitative examples of report generation. We present several examples with reports generated by RadFM and reference reports. The green color highlights accurate keywords, while the red color indicates prediction errors. Additionally, the blue color denotes instances where the model missed this information that has been mentioned in the reference reports.}
    \label{fig:result_report}
\end{figure}

\change{In Fig.~\ref{fig:result_report}, we provide qualitative examples of radiology reports generation task by RadFM. It can be observed that the model is capable of identifying the underlying diseases and, in some cases, performs exceptionally well. However, the report generated by RadFM may lack specific location information, such as the `left' or `right' of an anatomical region.}

\begin{figure}[!htb]
    \centering
    \includegraphics[width = \textwidth]{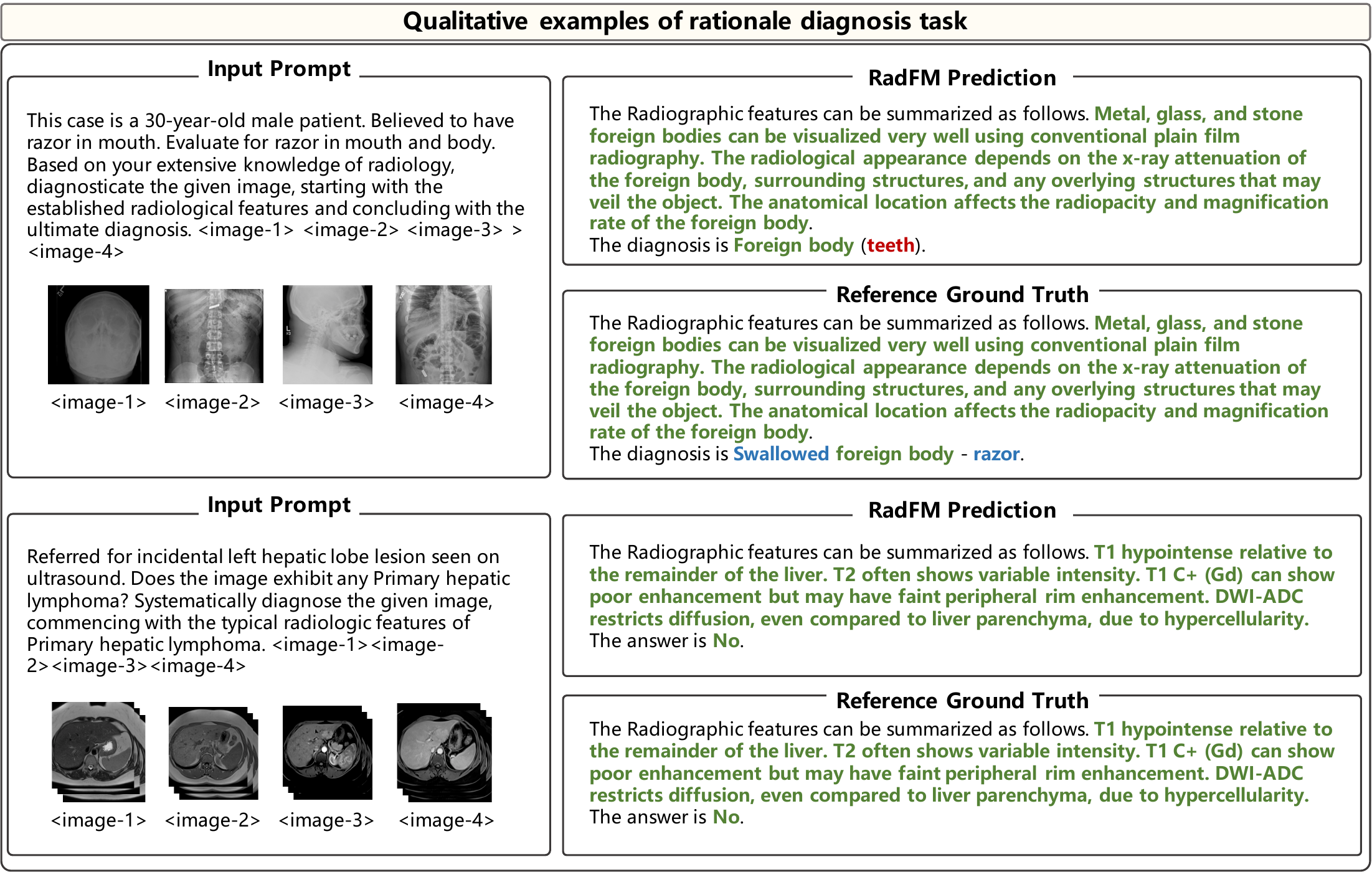}
    \vspace{4pt}
    \caption{Qualitative examples of rationale diagnosis. We present several examples generated by RadFM and reference ground truth. The green color highlights accurate keywords, while the red color indicates prediction errors. Additionally, the
blue color denotes instances where the model missed this information that has been mentioned in the reference ground truth.}
    \label{fig:result_rationale}
\end{figure}

\change{At last, Fig.~\ref{fig:result_rationale} shows two rationale diagnosis cases. 
The first case is a patient with pulmonary embolism and the latter is with subarachnoid haemorrhage. On both cases, RadFM can make accurate diagnosis in free form and give further related radiologic reasoning. 
However, the limitation can also be observed that the reasoning results are still general and more like background medical knowledge, yet not specific to the input case.}

\section{Discussion}
\subsection{RadFM is the first foundation model that unified 2D and 3D radiologic images.}
In the field of radiologic images, one significant challenge on developing foundation model lies on the disparity of image dimensions, 
{\em i.e.}, medical scans are either 2D or 3D, 
posing challenges on integrating real 3D MRI or CT images alongside with 2D images like X-rays or ultrasounds. 
As a consequence, the development of foundational models has been significantly impeded, with most current models only accommodating 2D images.
To overcome these limitations, 
we propose a new training structure that unifies 2D and 3D images, allowing a wide variety of real clinical images to be fed into one network.
By unifying the training process, 
our model benefits from a more comprehensive understanding of the diverse clinical images, leading to improved performance and versatility.
Additionally, to facilitate research and foster collaboration in the field, 
we collect a new medical multimodal dataset, 
containing 16M 2D or 3D medical scans with text descriptions or labels.

\subsection{RadFM unifies the medical tasks with a generative model.}
While developing AI for medicine, traditional approaches consider a divide-and-conquer idea, that tackles a myriad of specific tasks individually, 
such as diagnosis, report generation, and medical visual question answering, resulting in separated approaches with limited generalization ability and efficiency. 
Here, we formulate diverse medical tasks as multi-modal question-answering and develop a generative visual-language model RadFM that can answer arbitrary questions or follow instructions. 
\change{
Different from existing works with the use of exemplars in prompts, 
we use zero-shot prompt for all the tasks, 
allowing users to interact with the model without providing any exemplar images, questions, and answers.
Training models support zero-shot prompts is certainly more challenging, however, considering the user might be patients without with no clinical background, or examplar images, zero-shot prompt would be more desiable for real application. 
}
By unifying the tasks, RadFM achieves promising performance across a wide spectrum of clinical tasks. 
On the medical VQA task, RadFM surpasses the performance of MedVINT, a pre-trained model trained solely on a single Medical VQA dataset.

\subsection{RadFM supports multiple images as input.}
Till now, most existing multi-modal foundation models in the medical field have been limited to supporting only single image input per interaction. 
However, such design poses critical challenges in medical scenarios where diagnosis and treatment decisions often necessitate longitudinal clinical follow-ups, that comprise a series of radiologic images.
To overcome this limitation and pave the way for more comprehensive medical image analysis, our proposed RadFM supports multi-image input. 
To support the training, our constructed dataset is largely composed of multi-image input data, and our innovative training flow seamlessly accommodates this unique medical scenery, fostering advancements in medical image analysis.

\subsection{A general evaluation benchmark for radiology foundation models.} 
Evaluating the performance of medical foundation models is a complex undertaking due to the specialized nature of medical tasks. 
In the pursuit of advancing radiology foundation models, 
we propose RadBench, 
a novel benchmark that encompasses a diverse range of medical scenarios. 
The benchmark comprises 5 tasks, namely modality recognition, 
disease diagnosis, visual question answering, and report generation. 
By incorporating both 2D and 3D images, RadBench offers a more comprehensive and realistic evaluation platform compared to existing benchmarks. Additionally, as existing evaluation metrics are primarily designed for general natural language tasks, which may not adequately capture the intricacies and nuances specific to medical image analysis, thus may not reflect the model's true capabilities in real-world clinical scenarios. To address this limitation, we propose two new evaluation metrics, namely UMLS\_Precision and UMLS\_Recall. 
Unlike conventional metrics, UMLS Precision and Recall are tailored to measure the model's performance in medical tasks. 
By leveraging the Unified Medical Language System (UMLS), a comprehensive medical knowledge resource, these metrics provide a more tailored evaluation, ensuring that the model's outputs align with medical domain expertise.

\subsection{The superiority of RadFM.}
As shown in Tab.~\ref{tab_zeroshot_1} and Fig.\ref{fig:teaser}, 
while evaluating on our proposed comprehensive benchmark for radiology, namely, RadBench,
RadFM outperforms previous methods by a significant margin across all five tasks, 
showcasing its exceptional capabilities. 
Notably, RadFM excels in particularly challenging tasks such as medical VQA, report generation, and rationale diagnosis, which demand a profound understanding of both textual information and images. 
\change{The average human evaluation score for RadFM in these tasks surpasses that of GPT-4V, especially in the medical VQA task, where RadFM achieves a score of 2.87 compared to GPT-4V's score of 2.13.}
In medical VQA, the questions can be drastically varying,  
from simple queries like ``What modality is the given image?'' to more complex and context-rich questions, such as ``Based on the provided images, patient data (age, gender, medical history), can you identify the disease that is commonly associated with such radiological manifestations?'' 
The complexity of questions makes medical VQA a comprehensive and versatile task.
By integrating visual and textual information, RadFM can handle these varying question types, delivering accurate and meaningful answers. 
Similarly, in report generation, RadFM showcases significant improvement. 
The model's ability to discern relevant information from the provided images and weave it cohesively with textual prompts leads to highly informative and contextually rich reports, setting it apart from traditional methods. 
Overall, the performance of RadFM across these diverse tasks confirms its versatility and transformative potential in medical image analysis. 

\subsection{Clinical impact.}
\change{
In clinical practice, a reliable foundation model should satisfy the following key points: 
\vspace{-0.15cm}
\begin{itemize}
\setlength\itemsep{0.15cm}
    \item \textbf{Support 3D Data:} In practical clinical settings, 
    CT and MRI are widely used and the diagnosis of most diseases heavily relies on them. 
    Our model design enables to handle real clinical imaging data with one unified architecture,
    {\em e.g.}, 2D or 3D.
    \item \textbf{Multiple Scans:} The comprehensive diagnosis usually requires the input of multi-scans from various imaging modalities, or sometimes history radiologic images, our model's ability to digest multiple scans therefore meets such clinical demand.
    \item \textbf{Interleaved Data Format:} In clinical practice, images analysis often requires the knowledge of the patient's history or background. 
    The interleaved data format allows the user to freely input additional background information, ensuring the model can combine multi-source information to finish complex clinical decision-making tasks.
\end{itemize}
In contrast to all existing medical foundation models, 
RadFM is the first model that simultaneously satisfies the abovementioned criteria, {\em i.e.}, it allows users to input 3D multiple scans interleaved with texts per query, which can greatly benefit its clinical usage. }

\subsection{Limitations.}

Despite our efforts in developing a foundation model for radiology, RadFM still exhibits several limitations:

{\em First}, the capacity to generate meaningful and accurate long sentences remains underdeveloped, causing the foundation models still far from clinically useful. As demonstrated in Tab.~\ref{tab_zeroshot_1}, for rationale diagnosis and report generation, the quantitative results surpass previous works but are still far from practical satisfactory. In human rating, similar results are also observed. As shown in Fig.~\ref{fig:clinical_evaluation}, none of the model gets over score 3 that represents moderately accurate, showing that there is still long way to go for developing generalist medical foundation models.

{\em Second}, the proportion of actual 3D images in the data remains limited. As illustrated in Fig.\ref{fig:data_ana}, although we attempt to compensate for the lack of 3D images, yet 2D images remain to be dominating.

{\em Third}, the automatic evaluation metrics fall short of expectations. Compared to general contexts where the emphasis is placed on the overall coherence and fluency of sentences, medical texts prioritize precision in key statements and contain many synonyms, like `MRI' and `Magnetic Resonance Imaging', overlooking minor syntax errors. Although we employ UMLS\_Precision and UMLS\_Recall to mitigate this issue, they do not fully reflect true performance. On the other hand, though human evaluation is flexible and accurate, it is costly and cannot be carried out on a large scale. A robust automatic evaluation metric is essential to guide the construction of reliable and robust medical foundation models.

{\em Fourth}, as the 3D images in our dataset are downloaded from the internet, some metadata is missing, for example, the imaging spacing. Such lack of precise distance measurement makes it impossible to make certain statements, such as ``The tumor is 3cm large''. 
This specification is crucial for report writing. 

\change{
{\em Lastly}, due to the scale of dataset~(16M image-text pairs), model size~(14B parameters), investigating the effects of different components becomes increasingly challenging and prohibitively expensive in terms of both time and computational resources.
as future work, we will further break down the problem and investigate each component, ultimately enhancing our understanding and refining the model's performance.
}



\section{Related Work}
With the success of generative language foundation models such as GPT-4~\cite{OpenAI2023GPT4TR} and PaLM-2~\cite{Anil2023PaLM2T}, there has been a surge of interest in multi-modal foundation models. While significant strides have been made in the realm of natural scenery, as evidenced by BLIP-2~\cite{Li2023BLIP2BL} and Flamingo~\cite{Alayrac2022FlamingoAV}, the development of generalist medical artificial intelligence is still in its nascent stages \cite{moor2023foundation}. The relevant research can be bifurcated into two primary areas, namely, dataset construction and model training. 

\vspace{3pt} \noindent \textbf{Dataset Construction.} 
Contrary to the natural scenery domain, which boasts numerous large-scale multi-modal datasets such as MMC4~\cite{zhu2023multimodal}, Visual Genome~\cite{krishna2017visual}, and LION-5B~\cite{schuhmann2022laion}, the medical domain is somewhat lacking. The most widely utilized medical multi-modal dataset is MIMIC-CXR~\cite{johnson2019mimic}, which only contains chest X-ray images with caption reports and its quantity~(224K) is relatively small. In PMC-OA~\cite{lin2023pmc}, the authors have compiled a dataset containing 1.6M image-caption pairs. Although it encompasses various image modalities, many 3D medical scans are presented as 2D slices since the images are extracted from papers. There are also some medical VQA datasets, such as VQA-RAD~\cite{lau2018dataset}, SLAKE~\cite{liu2021slake}, and PMC-VQA~\cite{Zhang2023PMCVQAVI}, but they are also limited to 2D images. In Med-Flamingo~\cite{moor2023medflamingo}, they have collected a dataset, MTB, consisting of approximately 0.8M images interleaved with texts while it is not open-source.
~\change{
Consequently, due to these limitations in available data, existing medical foundation models have concentrated on a narrow range of data modalities.
For example, LLaVA-Med~\cite{li2023llava} and MedVInT~\cite{Zhang2023PMCVQAVI} utilize image captions in PubMed Central, which exists a significant domain gap between real-world clinical data. 
In Med-PaLM M~\cite{tu2023towards}, the authors amalgamate existing medical images or multi-modal datasets, but the majority of images are X-rays, which are not sufficiently accurate for clinical practice.
}

\vspace{3pt} \noindent \textbf{Model Training.} 
\change{
To date, several works have focused on building medical foundation models, yet most of these works are limited to support for 2D images~\cite{li2023llava, Zhang2023PMCVQAVI,tu2023towards,moor2023medflamingo}.
An ideal foundation model should exhibit a comprehensive set of capabilities: it should support both 2D and 3D image inputs, multi-image input per case, and images interleaved with text inputs.
Currently, there exists no model that can simultaneously support this series of heterogeneous input forms.
This paper aims to address these gaps, aligning more closely with clinical practice.
}

\section{Conclusion}
In conclusion, in this paper, we have constructed a complete set of medical foundation model-building processes, including data collection, problem formulation, model design, training, and evaluation. We construct the largest medical multi-modal database in this paper. In model capabilities, compared to existing work, our model is able to process multiple 3D or 2D image inputs interleaved with texts, which fits the practical usage more. We surpass the latest open-source multi-modal foundation model significantly.
We will release all corresponding data, codes, and models. We believe this can greatly promote the development of medical foundation models.

\section{Data availability}
The datasets used for pre-training are listed in Table~\ref{data_available}. 

\begin{table}[!htb]
\footnotesize
\centering
\renewcommand{\arraystretch}{1.2}
\caption{Data availability. For the Rad3D-series and MPx-series datasets, researchers may obtain by reasonable request from the authors. }
\vspace{3pt}
\begin{tabular}{lp{9cm}<{\raggedright}l}
\hline
Dataset Name & Link & Access \\ 
\hline
Rad3D-series & \change{We will release and remove restriction once being accepted.} & Restricted Access \\
MPx-series & \change{We will release and remove restriction once being accepted.} & Restricted Access \\
PMC-Inline & \url{https://huggingface.co/datasets/chaoyi-wu/PMC-Inline} & Open Access \\
PMC-CaseReport & \url{https://huggingface.co/datasets/chaoyi-wu/PMC-CaseReport_original} \url{https://huggingface.co/datasets/chaoyi-wu/PMC-CaseReport} & Open Access \\
VinDr-Mammo~\cite{vindrmammo} & \url{https://www.physionet.org/content/vindr-mammo/1.0.0/} & Credentialed Access \\
VinDr-SpineXR~\cite{vindrspine} & \url{https://www.physionet.org/content/vindr-spinexr/1.0.0/} & Credentialed Access \\
VinDr-PCXR~\cite{vindrpcxr} & \url{https://physionet.org/content/vindr-pcxr/1.0.0/} & Credentialed Access \\
RadChest-CT~\cite{anas_awadalla_2023_7733589} & \url{https://zenodo.org/record/6406114/} & Credentialed Access \\
PMC-OA~\cite{lin2023pmc} & \url{https://huggingface.co/datasets/axiong/pmc_oa_beta} & Open Access \\
PMC-VQA~\cite{Zhang2023PMCVQAVI} & \url{https://huggingface.co/datasets/xmcmic/PMC-VQA} & Open Access \\
VQA-RAD~\cite{lau2018dataset} & \url{https://osf.io/89kps/} & Open Access \\
SLAKE~\cite{liu2021slake} & \url{https://www.med-vqa.com/slake/} & Open Access \\
MIMIC-CXR~\cite{johnson2019mimic} & \url{https://physionet.org/content/mimic-cxr/2.0.0} & Credentialed Access \\
VinDr-CXR~\cite{nguyen2022vindr} & \url{https://physionet.org/content/vindr-cxr/1.0.0/} & Credentialed Access \\
NIH ChestXray14~\cite{wang2017chestx} & \url{https://nihcc.app.box.com/v/ChestXray-NIHCC/folder/36938765345} & Open Access \\
CheXpert~\cite{irvin2019chexpert} & \url{https://aimi.stanford.edu/chexpert-chest-x-rays} & Open Access \\
Covid-CXR2~\cite{pavlova2021covid} & \url{https://www.kaggle.com/datasets/andyczhao/covidx-cxr2}& Open Access \\
NLM-TB~\cite{jaeger2014two} & \url{https://openi.nlm.nih.gov/imgs/collections/NLM-MontgomeryCXRSet.zip}  \url{https://openi.nlm.nih.gov/imgs/collections/ChinaSet_AllFiles.zip} & Open Access \\
Object-CXR~\cite{objectcxr} & \url{https://web.archive.org/web/20201127235812/https://jfhealthcare.github.io/object-CXR/} & Open Access \\
OpenI~\cite{demner2016preparing} & \url{https://www.kaggle.com/datasets/raddar/chest-xrays-indiana-university} & Open Access \\
RSNA~\cite{shih2019augmenting}& \url{https://www.rsna.org/education/ai-resources-and-training/ai-image-challenge/rsna-pneumonia-detection-challenge-2018}& Open Access \\
SIIM-ACR~\cite{filice2020crowdsourcing} & \url{https://www.kaggle.com/datasets/jesperdramsch/siim-acr-pneumothorax-segmentation-data}& Open Access \\
\bottomrule
\end{tabular}
\label{data_available}
\end{table}

\section{CODE availability}
The code is available on GitHub at \url{https://github.com/chaoyi-wu/RadFM}.

\clearpage

\bibliographystyle{sn-mathphys} 
\bibliography{references} 

\begin{thebibliography}{10}\itemsep=-1pt

\bibitem{Alayrac2022FlamingoAV}
Jean-Baptiste Alayrac, Jeff Donahue, Pauline Luc, Antoine Miech, Iain Barr,
  Yana Hasson, Karel Lenc, Arthur Mensch, Katie Millican, Malcolm Reynolds,
  Roman Ring, Eliza Rutherford, Serkan Cabi, Tengda Han, Zhitao Gong, Sina
  Samangooei, Marianne Monteiro, Jacob Menick, Sebastian Borgeaud, Andy Brock,
  Aida Nematzadeh, Sahand Sharifzadeh, Mikolaj Binkowski, Ricardo Barreira,
  Oriol Vinyals, Andrew Zisserman, and Karen Simonyan.
\newblock Flamingo: a visual language model for few-shot learning.
\newblock In {\em Advances in Neural Information Processing Systems (NeurIPS)},
  2022.

\bibitem{ALFARGHALY2021100557}
Omar Alfarghaly, Rana Khaled, Abeer Elkorany, Maha Helal, and Aly Fahmy.
\newblock Automated radiology report generation using conditioned transformers.
\newblock {\em Informatics in Medicine Unlocked}, 24:100557, 2021.

\bibitem{Anil2023PaLM2T}
Rohan Anil, Andrew~M. Dai, Orhan Firat, Melvin Johnson, Dmitry Lepikhin,
  Alexandre~Tachard Passos, Siamak Shakeri, Emanuel Taropa, Paige Bailey, Z.
  Chen, Eric Chu, J. Clark, Laurent~El Shafey, Yanping Huang, Kathleen~S.
  Meier-Hellstern, Gaurav Mishra, Erica Moreira, Mark Omernick, Kevin Robinson,
  Sebastian Ruder, Yi Tay, Kefan Xiao, Yuanzhong Xu, Yujing Zhang,
  Gustavo~Hernandez Abrego, Junwhan Ahn, Jacob Austin, Paul Barham, Jan~A.
  Botha, James Bradbury, Siddhartha Brahma, Kevin~Michael Brooks, Michele
  Catasta, Yongzhou Cheng, Colin Cherry, Christopher~A. Choquette-Choo,
  Aakanksha Chowdhery, C Cr{\'e}py, Shachi Dave, Mostafa Dehghani, Sunipa Dev,
  Jacob Devlin, M.~C. D'iaz, Nan Du, Ethan Dyer, Vladimir Feinberg, Fan Feng,
  Vlad Fienber, Markus Freitag, Xavier Garc{\'i}a, Sebastian Gehrmann, Lucas
  Gonz{\'a}lez, Guy Gur-Ari, Steven Hand, Hadi Hashemi, Le Hou, Joshua Howland,
  An~Ren Hu, Jeffrey Hui, Jeremy Hurwitz, Michael Isard, Abe Ittycheriah,
  Matthew Jagielski, Wen~Hao Jia, Kathleen Kenealy, Maxim Krikun, Sneha
  Kudugunta, Chang Lan, Katherine Lee, Benjamin Lee, Eric Li, Mu-Li Li, Wei Li,
  Yaguang Li, Jun~Yu Li, Hyeontaek Lim, Han Lin, Zhong-Zhong Liu, Frederick
  Liu, Marcello Maggioni, Aroma Mahendru, Joshua Maynez, Vedant Misra, Maysam
  Moussalem, Zachary Nado, John Nham, Eric Ni, Andrew Nystrom, Alicia Parrish,
  Marie Pellat, Martin Polacek, Alex Polozov, Reiner Pope, Siyuan Qiao, Emily
  Reif, Bryan Richter, Parker Riley, Alexandra Ros, Aurko Roy, Brennan Saeta,
  Rajkumar Samuel, Renee~Marie Shelby, Ambrose Slone, Daniel Smilkov, David~R.
  So, Daniela Sohn, Simon Tokumine, Dasha Valter, Vijay Vasudevan, Kiran
  Vodrahalli, Xuezhi Wang, Pidong Wang, Zirui Wang, Tao Wang, John Wieting,
  Yuhuai Wu, Ke Xu, Yunhan Xu, Lin~Wu Xue, Pengcheng Yin, Jiahui Yu, Qiaoling
  Zhang, Steven Zheng, Ce Zheng, Wei Zhou, Denny Zhou, Slav Petrov, and Yonghui
  Wu.
\newblock Palm 2 technical report.
\newblock {\em ArXiv}, abs/2305.10403, 2023.

\bibitem{armato2011lung}
Samuel~G Armato~III, Geoffrey McLennan, Luc Bidaut, Michael~F McNitt-Gray,
  Charles~R Meyer, Anthony~P Reeves, Binsheng Zhao, Denise~R Aberle, Claudia~I
  Henschke, Eric~A Hoffman, et~al.
\newblock The lung image database consortium (lidc) and image database resource
  initiative (idri): a completed reference database of lung nodules on ct
  scans.
\newblock {\em Medical physics}, 38(2):915--931, 2011.

\bibitem{anas_awadalla_2023_7733589}
Anas Awadalla, Irena Gao, Joshua Gardner, Jack Hessel, Yusuf Hanafy, Wanrong
  Zhu, Kalyani Marathe, Yonatan Bitton, Samir Gadre, Jenia Jitsev, Simon
  Kornblith, Pang~Wei Koh, Gabriel Ilharco, Mitchell Wortsman, and Ludwig
  Schmidt.
\newblock Openflamingo, Mar. 2023.

\bibitem{bazi2023vision}
Yakoub Bazi, Mohamad Mahmoud~Al Rahhal, Laila Bashmal, and Mansour Zuair.
\newblock Vision--language model for visual question answering in medical
  imagery.
\newblock {\em Bioengineering}, 10(3):380, 2023.

\bibitem{bodenreider2004unified}
Olivier Bodenreider.
\newblock The unified medical language system (umls): integrating biomedical
  terminology.
\newblock {\em Nucleic acids research}, 32(suppl\_1):D267--D270, 2004.

\bibitem{bommasani2021opportunities}
Rishi Bommasani, Drew~A Hudson, Ehsan Adeli, Russ Altman, Simran Arora, Sydney
  von Arx, Michael~S Bernstein, Jeannette Bohg, Antoine Bosselut, Emma
  Brunskill, et~al.
\newblock On the opportunities and risks of foundation models.
\newblock {\em arXiv preprint arXiv:2108.07258}, 2021.

\bibitem{Bustos2019PadChestAL}
Aurelia Bustos, A. Pertusa, Josee~Mar{\'i}a Salinas, and Mar{\'i}a de~la
  Iglesia-Vay{\'a}.
\newblock Padchest: A large chest x-ray image dataset with multi-label
  annotated reports.
\newblock {\em Medical image analysis}, 66:101797, 2019.

\bibitem{chatterjee2022classification}
Soumick Chatterjee, Faraz~Ahmed Nizamani, Andreas N{\"u}rnberger, and Oliver
  Speck.
\newblock Classification of brain tumours in mr images using deep spatiospatial
  models.
\newblock {\em Scientific Reports}, 12(1):1505, 2022.

\bibitem{chen2016training}
Tianqi Chen, Bing Xu, Chiyuan Zhang, and Carlos Guestrin.
\newblock Training deep nets with sublinear memory cost.
\newblock {\em arXiv preprint arXiv:1604.06174}, 2016.

\bibitem{demner2016preparing}
Dina Demner-Fushman, Marc~D Kohli, Marc~B Rosenman, Sonya~E Shooshan, Laritza
  Rodriguez, Sameer Antani, George~R Thoma, and Clement~J McDonald.
\newblock Preparing a collection of radiology examinations for distribution and
  retrieval.
\newblock {\em Journal of the American Medical Informatics Association},
  23(2):304--310, 2016.

\bibitem{Dosovitskiy2020AnII}
Alexey Dosovitskiy, Lucas Beyer, Alexander Kolesnikov, Dirk Weissenborn,
  Xiaohua Zhai, Thomas Unterthiner, Mostafa Dehghani, Matthias Minderer, Georg
  Heigold, Sylvain Gelly, Jakob Uszkoreit, and Neil Houlsby.
\newblock An image is worth 16x16 words: Transformers for image recognition at
  scale.
\newblock {\em ArXiv}, abs/2010.11929, 2020.

\bibitem{draelos_rachel_lea_2020_6406114}
Rachel~Lea Draelos, David Dov, Maciej~A Mazurowski, Joseph~Y. Lo, Ricardo
  Henao, Geoffrey~D. Rubin, and Lawrence Carin.
\newblock Rad-chestct dataset, Oct. 2020.

\bibitem{filice2020crowdsourcing}
Ross~W Filice, Anouk Stein, Carol~C Wu, Veronica~A Arteaga, Stephen
  Borstelmann, Ramya Gaddikeri, et~al.
\newblock Crowdsourcing pneumothorax annotations using machine learning
  annotations on the nih chest x-ray dataset.
\newblock {\em Journal of digital imaging}, 33:490--496, 2020.

\bibitem{objectcxr}
JF Healthcare.
\newblock Object-cxr - automatic detection of foreign objects on chest x-rays.
\newblock 2020.

\bibitem{irvin2019chexpert}
Jeremy Irvin et~al.
\newblock Chexpert: A large chest radiograph dataset with uncertainty labels
  and expert comparison.
\newblock In {\em Proceedings of the AAAI conference on artificial
  intelligence}, volume~33, pages 590--597, 2019.

\bibitem{jaeger2014two}
Stefan Jaeger, Sema Candemir, Sameer Antani, Y{\`\i}-Xi{\'a}ng~J W{\'a}ng,
  Pu-Xuan Lu, and George Thoma.
\newblock Two public chest x-ray datasets for computer-aided screening of
  pulmonary diseases.
\newblock {\em Quantitative imaging in medicine and surgery}, 4(6):475, 2014.

\bibitem{jaegle2021perceiver}
Andrew Jaegle, Felix Gimeno, Andy Brock, Oriol Vinyals, Andrew Zisserman, and
  Joao Carreira.
\newblock Perceiver: General perception with iterative attention.
\newblock In {\em International conference on machine learning}, pages
  4651--4664. PMLR, 2021.

\bibitem{johnson2019mimic}
Alistair~EW Johnson, Tom~J Pollard, Seth~J Berkowitz, Nathaniel~R Greenbaum,
  Matthew~P Lungren, Chih-ying Deng, Roger~G Mark, and Steven Horng.
\newblock Mimic-cxr, a de-identified publicly available database of chest
  radiographs with free-text reports.
\newblock {\em Scientific data}, 6(1):317, 2019.

\bibitem{krishna2017visual}
Ranjay Krishna, Yuke Zhu, Oliver Groth, Justin Johnson, Kenji Hata, Joshua
  Kravitz, Stephanie Chen, Yannis Kalantidis, Li-Jia Li, David~A Shamma, et~al.
\newblock Visual genome: Connecting language and vision using crowdsourced
  dense image annotations.
\newblock {\em International journal of computer vision}, 123:32--73, 2017.

\bibitem{lau2018dataset}
Jason~J Lau, Soumya Gayen, Asma Ben~Abacha, and Dina Demner-Fushman.
\newblock A dataset of clinically generated visual questions and answers about
  radiology images.
\newblock {\em Scientific data}, 5(1):1--10, 2018.

\bibitem{li2023llava}
Chunyuan Li, Cliff Wong, Sheng Zhang, Naoto Usuyama, Haotian Liu, Jianwei Yang,
  Tristan Naumann, Hoifung Poon, and Jianfeng Gao.
\newblock Llava-med: Training a large language-and-vision assistant for
  biomedicine in one day.
\newblock {\em arXiv preprint arXiv:2306.00890}, 2023.

\bibitem{Li2023BLIP2BL}
Junnan Li, Dongxu Li, Silvio Savarese, and Steven Hoi.
\newblock Blip-2: Bootstrapping language-image pre-training with frozen image
  encoders and large language models.
\newblock {\em ArXiv}, abs/2301.12597, 2023.

\bibitem{lin2004rouge}
Chin-Yew Lin.
\newblock Rouge: A package for automatic evaluation of summaries.
\newblock In {\em Text summarization branches out}, pages 74--81, 2004.

\bibitem{lin2023pmc}
Weixiong Lin, Ziheng Zhao, Xiaoman Zhang, Chaoyi Wu, Ya Zhang, Yanfeng Wang,
  and Weidi Xie.
\newblock Pmc-clip: Contrastive language-image pre-training using biomedical
  documents.
\newblock In {\em Medical Image Computing and Computer Assisted Intervention
  (MICCAI) 2023}. Springer, 2023.

\bibitem{liu2021slake}
Bo Liu, Li-Ming Zhan, Li Xu, Lin Ma, Yan Yang, and Xiao-Ming Wu.
\newblock Slake: A semantically-labeled knowledge-enhanced dataset for medical
  visual question answering.
\newblock In {\em 2021 IEEE 18th International Symposium on Biomedical Imaging
  (ISBI)}, pages 1650--1654. IEEE, 2021.

\bibitem{monshi2020deep}
Maram Mahmoud~A Monshi, Josiah Poon, and Vera Chung.
\newblock Deep learning in generating radiology reports: A survey.
\newblock {\em Artificial Intelligence in Medicine}, 106:101878, 2020.

\bibitem{Moor2023FoundationMF}
Michael Moor, Oishi Banerjee, Zahra F~H Abad, Harlan~M. Krumholz, Jure
  Leskovec, Eric~J. Topol, and Pranav Rajpurkar.
\newblock Foundation models for generalist medical artificial intelligence.
\newblock {\em Nature}, 616:259--265, 2023.

\bibitem{moor2023foundation}
Michael Moor, Oishi Banerjee, Zahra Shakeri~Hossein Abad, Harlan~M Krumholz,
  Jure Leskovec, Eric~J Topol, and Pranav Rajpurkar.
\newblock Foundation models for generalist medical artificial intelligence.
\newblock {\em Nature}, 616(7956):259--265, 2023.

\bibitem{moor2023medflamingo}
Michael Moor, Qian Huang, Shirley Wu, Michihiro Yasunaga, Cyril Zakka, Yash
  Dalmia, Eduardo~Pontes Reis, Pranav Rajpurkar, and Jure Leskovec.
\newblock Med-flamingo: A multimodal medical few-shot learner.
\newblock July 2023.
\newblock arXiv:2307.15189.

\bibitem{nguyen2022vindr}
Ha~Q Nguyen, Khanh Lam, Linh~T Le, Hieu~H Pham, Dat~Q Tran, Dung~B Nguyen,
  et~al.
\newblock Vindr-cxr: An open dataset of chest x-rays with radiologist’s
  annotations.
\newblock {\em Scientific Data}, 9(1):429, 2022.

\bibitem{vindrmammo}
Hieu~T Nguyen, Ha~Q Nguyen, Hieu~H Pham, Khanh Lam, Linh~T Le, Minh Dao, and
  Van Vu.
\newblock Vindr-mammo: A large-scale benchmark dataset for computer-aided
  diagnosis in full-field digital mammography.
\newblock {\em Scientific Data}, 10(1):277, 2023.

\bibitem{vindrspine}
Hieu~T Nguyen, Hieu~H Pham, Nghia~T Nguyen, Ha~Q Nguyen, Thang~Q Huynh, Minh
  Dao, and Van Vu.
\newblock Vindr-spinexr: A deep learning framework for spinal lesions detection
  and classification from radiographs.
\newblock In {\em Medical Image Computing and Computer Assisted
  Intervention--MICCAI 2021: 24th International Conference, Strasbourg, France,
  September 27--October 1, 2021, Proceedings, Part V 24}, pages 291--301.
  Springer, 2021.

\bibitem{vindrpcxr}
Ngoc~H Nguyen, Hieu~H Pham, Thanh~T Tran, Tuan~NM Nguyen, and Ha~Q Nguyen.
\newblock Vindr-pcxr: An open, large-scale chest radiograph dataset for
  interpretation of common thoracic diseases in children.
\newblock {\em medRxiv}, pages 2022--03, 2022.

\bibitem{OpenAI2023GPT4TR}
OpenAI.
\newblock Gpt-4 technical report.
\newblock {\em ArXiv}, abs/2303.08774, 2023.

\bibitem{GPT4V}
OpenAI(2023).
\newblock Chatgpt can now see, hear, and speak.
\newblock {\em https://openai.com/blog/chatgpt-can-now-see-hear-and-speak},
  2023.

\bibitem{papineni2002bleu}
Kishore Papineni, Salim Roukos, Todd Ward, and Wei-Jing Zhu.
\newblock Bleu: a method for automatic evaluation of machine translation.
\newblock In {\em Proceedings of the 40th annual meeting of the Association for
  Computational Linguistics}, pages 311--318, 2002.

\bibitem{pavlova2021covid}
Maya Pavlova et~al.
\newblock Covid-net cxr-2: An enhanced deep convolutional neural network design
  for detection of covid-19 cases from chest x-ray images.
\newblock {\em Frontiers in Medicine}, 9, 2022.

\bibitem{schuhmann2022laion}
Christoph Schuhmann, Romain Beaumont, Richard Vencu, Cade Gordon, Ross
  Wightman, Mehdi Cherti, Theo Coombes, Aarush Katta, Clayton Mullis, Mitchell
  Wortsman, et~al.
\newblock Laion-5b: An open large-scale dataset for training next generation
  image-text models.
\newblock {\em Advances in Neural Information Processing Systems},
  35:25278--25294, 2022.

\bibitem{shih2019augmenting}
George Shih, Carol~C Wu, Safwan~S Halabi, Marc~D Kohli, Luciano~M Prevedello,
  et~al.
\newblock Augmenting the national institutes of health chest radiograph dataset
  with expert annotations of possible pneumonia.
\newblock {\em Radiology: Artificial Intelligence}, 1(1):e180041, 2019.

\bibitem{tiu2022expert}
Ekin Tiu, Ellie Talius, Pujan Patel, Curtis~P Langlotz, Andrew~Y Ng, and Pranav
  Rajpurkar.
\newblock Expert-level detection of pathologies from unannotated chest x-ray
  images via self-supervised learning.
\newblock {\em Nature Biomedical Engineering}, 6(12):1399--1406, 2022.

\bibitem{touvron2023llama}
Hugo Touvron, Thibaut Lavril, Gautier Izacard, Xavier Martinet, Marie-Anne
  Lachaux, Timoth{\'e}e Lacroix, Baptiste Rozi{\`e}re, Naman Goyal, Eric
  Hambro, Faisal Azhar, et~al.
\newblock Llama: Open and efficient foundation language models.
\newblock {\em arXiv preprint arXiv:2302.13971}, 2023.

\bibitem{tu2023towards}
Tao Tu, Shekoofeh Azizi, Danny Driess, Mike Schaekermann, Mohamed Amin,
  Pi-Chuan Chang, Andrew Carroll, Chuck Lau, Ryutaro Tanno, Ira Ktena, et~al.
\newblock Towards generalist biomedical ai.
\newblock {\em arXiv preprint arXiv:2307.14334}, 2023.

\bibitem{van2023open}
Tom van Sonsbeek, Mohammad~Mahdi Derakhshani, Ivona Najdenkoska, Cees~GM Snoek,
  and Marcel Worring.
\newblock Open-ended medical visual question answering through prefix tuning of
  language models.
\newblock {\em arXiv preprint arXiv:2303.05977}, 2023.

\bibitem{wang2017chestx}
Xiaosong Wang, Yifan Peng, Le Lu, Zhiyong Lu, Mohammadhadi Bagheri, and
  Ronald~M Summers.
\newblock Chestx-ray8: Hospital-scale chest x-ray database and benchmarks on
  weakly-supervised classification and localization of common thorax diseases.
\newblock In {\em Proceedings of the IEEE conference on computer vision and
  pattern recognition}, pages 2097--2106, 2017.

\bibitem{wantlin2023benchmd}
Kathryn Wantlin, Chenwei Wu, Shih-Cheng Huang, Oishi Banerjee, Farah Dadabhoy,
  Veeral~Vipin Mehta, Ryan~Wonhee Han, Fang Cao, Raja~R. Narayan, Errol Colak,
  Adewole Adamson, Laura Heacock, Geoffrey~H. Tison, Alex Tamkin, and Pranav
  Rajpurkar.
\newblock Benchmd: A benchmark for modality-agnostic learning on medical images
  and sensors, 2023.

\bibitem{Wu2023KDiagKD}
Chaoyi Wu, Xiaoman Zhang, Yanfeng Wang, Ya Zhang, and Weidi Xie.
\newblock K-diag: Knowledge-enhanced disease diagnosis in radiographic imaging.
\newblock {\em MICCAI Workshop}, 2023.

\bibitem{Wu2023MedKLIPMK}
Chaoyi Wu, Xiaoman Zhang, Ya Zhang, Yanfeng Wang, and Weidi Xie.
\newblock Medklip: Medical knowledge enhanced language-image pre-training.
\newblock In {\em Proceedings of the IEEE International Conference on Computer
  Vision (ICCV)}, 2023.

\bibitem{wu2023pmc}
Chaoyi Wu, Xiaoman Zhang, Ya Zhang, Yanfeng Wang, and Weidi Xie.
\newblock Pmc-llama: Further finetuning llama on medical papers.
\newblock {\em arXiv preprint arXiv:2304.14454}, 2023.

\bibitem{yu2022evaluating}
Feiyang Yu, Mark Endo, Rayan Krishnan, Ian Pan, Andy Tsai, Eduardo~Pontes Reis,
  Eduardo Kaiser Ururahy~Nunes Fonseca, Henrique Min~Ho Lee, Zahra
  Shakeri~Hossein Abad, Andrew~Y Ng, et~al.
\newblock Evaluating progress in automatic chest x-ray radiology report
  generation.
\newblock {\em medRxiv}, pages 2022--08, 2022.

\bibitem{yu2023evaluating}
Feiyang Yu, Mark Endo, Rayan Krishnan, Ian Pan, Andy Tsai, Eduardo~Pontes Reis,
  Eduardo Kaiser Ururahy~Nunes Fonseca, Henrique Min~Ho Lee, Zahra
  Shakeri~Hossein Abad, Andrew~Y Ng, et~al.
\newblock Evaluating progress in automatic chest x-ray radiology report
  generation.
\newblock {\em Patterns}, 4(9), 2023.

\bibitem{zhang2019bertscore}
Tianyi Zhang, Varsha Kishore, Felix Wu, Kilian~Q Weinberger, and Yoav Artzi.
\newblock Bertscore: Evaluating text generation with bert.
\newblock {\em arXiv preprint arXiv:1904.09675}, 2019.

\bibitem{zhang2023knowledge}
Xiaoman Zhang, Chaoyi Wu, Ya Zhang, Yanfeng Wang, and Weidi Xie.
\newblock Knowledge-enhanced pre-training for auto-diagnosis of chest radiology
  images.
\newblock {\em Nature Communication}, 2023.

\bibitem{Zhang2023PMCVQAVI}
Xiaoman Zhang, Chaoyi Wu, Ziheng Zhao, Weixiong Lin, Ya Zhang, Yanfeng Wang,
  and Weidi Xie.
\newblock Pmc-vqa: Visual instruction tuning for medical visual question
  answering.
\newblock {\em ArXiv}, abs/2305.10415, 2023.

\bibitem{zhao2023pytorch}
Yanli Zhao, Andrew Gu, Rohan Varma, Liang Luo, Chien-Chin Huang, Min Xu, Less
  Wright, Hamid Shojanazeri, Myle Ott, Sam Shleifer, et~al.
\newblock Pytorch fsdp: experiences on scaling fully sharded data parallel.
\newblock {\em arXiv preprint arXiv:2304.11277}, 2023.

\bibitem{zhu2023multimodal}
Wanrong Zhu, Jack Hessel, Anas Awadalla, Samir~Yitzhak Gadre, Jesse Dodge, Alex
  Fang, Youngjae Yu, Ludwig Schmidt, William~Yang Wang, and Yejin Choi.
\newblock {Multimodal C4}: An open, billion-scale corpus of images interleaved
  with text.
\newblock {\em arXiv preprint arXiv:2304.06939}, 2023.

\end{thebibliography}


\end{document}